\documentclass{article}
\usepackage{silence}
\usepackage[table,dvipsnames]{xcolor}
\definecolor{mydarkblue}{rgb}{0,0.08,0.45}
\definecolor{note_fontcolor}{rgb}{0.800781, 0.800781, 0.800781}

    \usepackage[accepted]{icml2021}

\usepackage[utf8]{inputenc} %
\usepackage[T1]{fontenc}    %
\usepackage[hyphens]{url}            %
\usepackage[
    colorlinks,
    bookmarks=true,
    breaklinks=true,
    linkcolor=mydarkblue,
    citecolor=mydarkblue,
    filecolor=mydarkblue,
    urlcolor=mydarkblue,]{hyperref}       %
\usepackage{booktabs}       %
\usepackage{amsfonts}       %
\usepackage{nicefrac}       %
\usepackage{microtype}      %
\usepackage{multicol}
\usepackage{subcaption}
\usepackage{amssymb, amsmath, amsthm}
\usepackage{thmtools}
\usepackage{mathtools}
\usepackage{tcolorbox}

\usepackage{subfiles}

\usepackage{wrapfig}

\usepackage[inline]{enumitem}
\setlist[description]{%
    font={\normalfont\itshape},
}
\usepackage{breakurl}
\usepackage[capitalize,nameinlink]{cleveref}

\renewcommand{\algorithmicrequire}{\textbf{Input:}}
\renewcommand{\algorithmicensure}{\textbf{Output:}}

\newtheorem{thm}{Theorem}[section]

\newtheorem{assm}[thm]{Assumption}

\newtheorem{setup}[thm]{Setup}

\newtheorem{prop}[thm]{Proposition}
\theoremstyle{definition}
\newtheorem{defn}[thm]{Definition}
\theoremstyle{remark}
\newtheorem{rem}[thm]{Remark}

\renewcommand{\[}{\begin{eqnarray}}
\renewcommand{\]}{\end{eqnarray}}

\newcommand{\X}{\mathcal{\mathbf{X}}}
\newcommand{\Y}{\mathcal{\mathbf{Y}}}

\newcommand{\phantomeq}{\phantom{{}={}}}

\global\long\def\EV{\operatorname*{\mathbb{E}}}%
\global\long\def\asto{\xrightarrow{\mathrm{\mathrm{a.s.}}}}%

\global\long\def\Gaus{\mathcal{N}}%

\global\long\def\R{\mathbb{R}}%

\global\long\def\trsp{\top}%

\global\long\def\f#1#2{\frac{#1}{#2}}%

\newcommand{\odefeq}{\mathbin{\overset{\text{\tiny{def}}}{=}}}

\newcommand{\defeq}{\mathrel{\raisebox{-0.3ex}{$\odefeq$}}}

\global\long\def\disteq{\overset{\mathrm{d}}{=}}%

\global\long\def\Cov{\mathrm{Cov}}%

\newcommand{\refNonlinp}{\textsc{\ref{instr:nonlin+}}}
\newcommand{\refMatmul}{\textsc{\ref{instr:matmul}}}
\newcommand{\refNonlin}{\textsc{\ref{instr:nonlin}}}
\newcommand{\refMoment}{\textsc{\ref{instr:moment}}}
\newcommand{\netsort}{\texorpdfstring{{$\textsc{Netsor}\trsp$}}{NetsorT}}

\newcommand{\netsortp}{\texorpdfstring{{$\textsc{Netsor}\top^+$}}{NetsorT+}}

\makeatletter
\let\orgdescriptionlabel\descriptionlabel
\newcommand*{\@restrictlabeltext}[1]{#1\protected@edef\@currentlabel{#1}}
\newcommand*{\nolabel}[1]{#1}%
\renewcommand*{\descriptionlabel}[1]{%
  \let\orglabel\label
  \let\label\@gobble
  \let\orig@hfil\hfil
  \def\hfil{}%
  \let\nolabel\@gobble
  \let\restrictlabeltext\@firstofone
  \phantomsection
  \protected@edef\@currentlabel{#1}%
  \let\hfil\orig@hfil
  \let\label\orglabel
  \let\restrictlabeltext\@restrictlabeltext
  \orgdescriptionlabel{#1}%
}
\makeatother

\icmltitlerunning{Architectural Universality of Neural Tangent Kernel Training Dynamics}

\begin{document}
\twocolumn[
\icmltitle{
    Tensor Programs IIb:\\
    Architectural Universality of Neural Tangent Kernel Training Dynamics}

\icmlsetsymbol{equal}{*}

\begin{icmlauthorlist}
\icmlauthor{Greg Yang}{msr,equal}
\icmlauthor{Etai Littwin}{apple,equal}
\end{icmlauthorlist}

\icmlaffiliation{msr}{Microsoft Research}
\icmlaffiliation{apple}{Apple Research}

\icmlcorrespondingauthor{Greg Yang}{gregyang@microsoft.com}
\icmlcorrespondingauthor{Etai Littwin}{elittwin@apple.com}

\icmlkeywords{Machine Learning, ICML}

\vskip 0.3in
]

\printAffiliationsAndNotice{\icmlEqualContribution}

\begin{abstract}

\citet{yang2} recently showed that the Neural Tangent Kernel (NTK) at initialization has an infinite-width limit for a large class of architectures including modern staples such as ResNet and Transformers.
However, their analysis does not apply to training.
Here, we show the same neural networks (in the so-called \emph{NTK parametrization}) during training follow a kernel gradient descent dynamics in function space, where the kernel is the infinite-width NTK.
This completes the proof of the \emph{architectural universality} of NTK behavior.
To achieve this result, we apply the \emph{Tensor Programs} technique:
Write the entire SGD dynamics inside a Tensor Program and analyze it via the \emph{Master Theorem}.
To facilitate this proof, we develop a graphical notation for Tensor Programs.
\end{abstract}

\vspace{-2em}

\enlargethispage{\baselineskip}

\section{Introduction}
\cite{NTK}'s pioneering work showed that a multi-layer perceptron (MLP) trained by gradient descent (GD) evolves like a linear model.
This spurred a flurry of research papers using this insight to tackle the core questions in deep learning theory, from optimization to generalization in both finite and infinite width regimes.
\cite{NTK}'s argument consists of two observations:
\begin{description}
    \item[\textsc{ntkInit}\label{NTKinit}] For the output of a network $f(\xi;w)$ with parameters $w$ given example $\xi$, \cite{NTK} identified the kernel $\mathcal{K}(\xi,\bar{\xi}) = \langle \nabla f(\xi;w),\nabla f(\bar{\xi},w) \rangle$, known as the \emph{Neural Tangent Kernel (NTK)}.
    They showed that if $f$ is parametrized and initialized appropriately, then $\mathcal{K}$ converges to a deterministic kernel $\mathring{\mathcal{K}}$ as the width of the network tends to infinity.
    \item[\textsc{ntkTrain}\label{NTKtrain}] As the infinitely wide network is trained by gradient descent, the NTK remains frozen in its initial state, and the network evolves as by kernel gradient descent with kernel $\mathring{\mathcal K}$
\end{description}

In \cite{yang2}, the \ref{NTKinit} property was proven to hold for \emph{standard architectures}, meaning any composition of MLPs, recurrent neural networks (RNN), LSTMs \cite{lstm}, gated recurrent unit (GRU) \cite{gru}, convolutions \cite{conv2,conv1,conv3,conv4,conv5}, residual connections \cite{res1,res2}, batch normalization \cite{bn}, graph neural networks \cite{graph1,graph2,graph3,graph4,graph5} and attention \cite{attention1,attention2}, along with arbitrary weight sharing between components.
More generally, it holds for any architecture expressible in a so-called \emph{Tensor Program} \cite{yangScalingLimit,yang,yang2,yang3}, of which the standard architectures are a subset.
However, their reasoning is limited to initialization only.

A statement is \emph{architecturally universal} if it holds for any \emph{reasonable} neural architecture.
This is an informal property, but here we will formalize it by taking \emph{reasonable} to be ``expressable in Tensor Programs.''
By the expressiveness of such programs \citep{yang,yang2}, architectural universality is a fairly robust notion that covers present (and, we expect, future) architectures comprehensively.
In this terminology, \citep{yang2} showed that \ref{NTKinit} is architecturally universal.

\paragraph{Our Contribution}
We show the architectural universality of the entire NTK theory by proving \ref{NTKtrain} for the same architectures discussed above, including all standard architectures.
In the process, we introduce a new graphical form of Tensor Programs that is both required in our proofs and useful for the pedagogy of Tensor Programs.

\paragraph{The Tensor Program Series} This paper follows \cite{yangScalingLimit,yang,yang2,yang3,yang4} in the series.
While we number this paper ``IIb'' right after \cite{yang2}, we actually need the complete theoretical foundation developed in III \cite{yang3}.
See \cref{footnote:relyonTP3} for more details.

\section{Background}\label{sec:3}
Let $f(\xi;w) \in \mathbb{R}$ denote the (scalar) output of a neural network parameterized by $w$, given example $\xi$. To understand how the output changes with a slight change in the network parameters $w_0 - \delta w$, we may naively expand the network function using the first order Taylor expansion around a base point $w_0$:
\begin{align}
    f(\xi,w_0 - \delta w) - f(\xi;w_0) \approx \langle\nabla_wf(\xi,w_0),\delta w\rangle.
    \label{eqn:FOTaylor}
\end{align}
Under the SGD algorithm, the weight update $\delta w$ is given by the gradient $\delta w = -\eta \chi(\hat{\xi})\nabla_wf(\hat{\xi};w_0)$ where $\chi(\hat{\xi})$ is the loss derivative, $\hat{\xi}$ is a sample from the training set, and $\eta$ is the learning rate.
Plugging into \cref{eqn:FOTaylor}, we get:
\begin{align}\label{ntk}
    f(\xi,w_0 - \delta w) - f(\xi;w_0) \approx -\eta \chi(\hat{\xi})\mathcal{K}(\xi,\hat{\xi}).
\end{align}
where $\mathcal{K}(\xi,\hat{\xi}) = \langle\nabla_wf(\xi;w_0),\nabla_wf(\hat{\xi};w_0)\rangle$ is the NTK. 
The NTK theory of infinitely wide neural networks as first proposed by \cite{NTK} boils down to the the following observations:
\emph{When the width of $f$ tend to infinity, the NTK $\mathcal K$ converges to a fixed kernel $\mathring{\mathcal K}$ at random initialization, independent of the specific instantiation of the weights,
and remains frozen during the optimization process.}
\cref{ntk} then gives an accurate description of the output evolution with if we substitue $\mathcal K$ with $\mathring{\mathcal{K}}$.
The seemingly complex optimization trajectory of SGD therefore reduce to the convex trajectory of kernel gradient descent with a time-independent kernel $\mathring{\mathcal{K}}$.\\
Consider the output of the network $f\in \mathbb{R}^D$ on the full training dataset. As shown in \cite{NTK}, when the $L2$ loss is used the evolution of the output $f_t$ at time $t$ under continuous time GD (i.e.\ gradient flow) takes a simple form:
\begin{align*}
    f_t - f^\star = e^{-\eta \mathring{\mathcal{K}} t}(f_0 - f^\star).
\end{align*}
where $\mathring{\mathcal{K}}\in \mathbb{R}^{D\times D}$ is the full NTK matrix evaluated on the training data, $f^\star$ is the label function,  and $f_0$ is the output at initialization.
Hence, provided $\mathring{\mathcal{K}}$ is full rank, as $t \to \infty$ we have that $f_t \to f^\star$, and the network can fit the training data perfectly.  

\paragraph{Previous Approaches vs Ours}
A common theme in showing \ref{NTKtrain} for MLP is to derive high-probability bounds on the deviation of the NTK $\mathcal K$ from its initial value after training
(e.g. \citet{allen-zhu_convergence_2018,du_gradient_2018,zou_stochastic_2018}).%
\footnote{
In the original NTK paper \citep{NTK}, the limit is taken as each layer width goes to infinity sequentially, which already doesn't make sense for weight-tied architectures like RNNs.
}
Obtaining these bounds usually requires developing ad hoc methods on a per-architecture basis, hindering the scalability of the method to other settings.
In the present work we take a more holistic approach, leveraging the recently developed Tensor Programs framework \cite{yangScalingLimit,yang,yang2,yang3}.
It consists of two layers of arguments:
1) The bottom layer analyzes how the \emph{distribution of (pre-)activations} change throughout the course of training;
this crucially leverages the mathematical machinery of the Tensor Programs Master Theorem.%
\footnote{In particular, we need to use the Master Theorem in \citep{yang3}, so \citep{yang2} could not have obtained \ref{NTKtrain} at the same time as \ref{NTKinit}.}
2) The top layer packages these insights systematically via the notion of \emph{paths} so as to apply to any architecture expressible by a Tensor Program.
We will illustrate 1) through examples in \cref{sec:examples} and 2) through figures in \cref{sec:proofsketch}.

\paragraph{Setup and Notations}\label{notations}
In this paper, we will consider the architecture (including depth), data, and training time to be fixed as width $n\to\infty$.%
\footnote{They will affect the rate of convergence to the infinite-width limit, but since we are only concerned with whether convergence occurs, they do not appear in our theorem statements here.}
We describe common notations used in the remainder of the paper.
For simplicity, we will consider SGD with batch size 1 and learning rate $\eta$ (often set to 1 WLOG).%
\footnote{This generalizes readily to any batch size and learning rate.}
We use $\xi_t$ to denote the input and $\mathcal L_t$ to denote the loss function (absorbing the label) at step $t$.
More generally, subscript $t$ on any symbol means \emph{time $t$}.
However, for brevity, we abuse notation and shorthand $f_t$ for $f_t(\xi_t)$, and, for any (pre-)activation $x$, $x_t$ for $x_t(\xi_t)$.%
\footnote{We will not refer to the \emph{function} $x_t: \xi \to x_t(\xi)$ (likewise for $f_t, \chi_t$), so this abuse of notation should cause no confusion.}
We will also write $\chi_t$ for the loss derivative $\mathcal L_t'(f_t)$.
For any vector $x(\xi)$ we define $\delta x_{t+1}(\xi) \defeq \sqrt{n}\big(x_{t+1}(\xi) - x_t(\xi)\big)$ and $dx(\xi) \defeq \sqrt{n}\frac{\partial f(\xi)}{\partial x(\xi)}$.
We will track the evolution of $f$ on an arbitrary input $\tilde \xi$.%
\footnote{It might help to think of $\tilde \xi$ as some \emph{test sample}, but it can also fall in the training set.}
Similar to above, we shorthand $\tilde{x}_t,\tilde{f}_t$ for $x_t(\tilde \xi), f_t(\tilde x)$.

\section{Motivating Examples}\label{sec:examples}
The purpose of this section is to illustrate our key ideas via simple, intuitive examples without diving into the specifics of Tensor Programs.
In the process, we will gain insight into how randomness from initialization propagates over the course of training.
As these examples intend to provide the reader with the proper intuition, we use informal arguments alone and relegate all formal statements to the appendix.
For brevity, we will gloss over minor details or routine calculations, but interested readers can see \cref{sec:appebdix_extra} for these omissions.

\paragraph{Key Idea}
It turns out that the random initialization and the overparametrization of weights cause each (pre-)activation vector $x_t(\xi)\in \R^n$, its gradient $dx_t(\xi)\in \R^n$, and its (scaled) change $\delta x_t(\xi) \in \R^n$ every time step $t$ to have roughly iid coordinates, not just initially but \emph{throughout training}.%
\footnote{This is a consequence of the Tensor Program Master Theorem.}
Then, as we shall demonstrate through the examples below, to track the evolution of the neural network function, it suffices to track the evolution of the coordinate distributions of $x(\xi)$, $dx(\xi)$, $\delta x(\xi)$.
We write $Z^{x(\xi)}$, $Z^{dx(\xi)}$, $Z^{\delta x(\xi)} \in \mathbb{R}$ for the random variables corresponding to such coordinate distributions.%
\footnote{As we will explain below, different $Z$s may correlate, reflecting correlations between corresponding vectors.}

Our goal is to derive, from these insights,
\begin{restatable}{claim}{claimNTK}
    \label{claim:NTK}
In the large width limit, $\tilde f_t = f_t(\tilde \xi)$ changes by
\begin{align}\label{kernel_d_1}
    \lim_{n\to\infty} \tilde{f}_{t+1} - \tilde{f}_t = -\mathring\chi_t \mathring{\mathcal{K}}(\tilde{\xi},\xi_t)
\end{align}
at step $t$, where $\mathring{\mathcal{K}}$ is the limiting NTK of the architecture and $\mathring\chi_t = \mathcal L_t'(\lim_{n} f_t)$ is the loss derivative.
\end{restatable}

We start with an example derivation for 1-hidden-layer MLP, before moving on to 2-hidden-layers, where the mathematics quickly become much more involved.

\subsection{1 Hidden Layer}

\label{sec:1LP}

Consider a 1-hidden-layer network with nonlinearity $\phi$:
\begin{align*}
f = V^\top x ,~~~x = \phi(h) ,~~~ h = U\xi
\end{align*}
where $\xi \in \mathbb{R}^d,~ U = \frac{u}{\sqrt{d}} \in \mathbb{R}^{n \times d},~V = \frac{v}{\sqrt{n}} \in \mathbb{R}^{n \times 1}$, for trainable parameter tensor $u$, initialized iid from $\Gaus(0,1)$.
In the interest of clarity we assume the output layer is not trained, and $d=1$.

For a vector $v \in \R^n$, let $v = \Theta(n^{a})$ mean that ``$v$ has coordinates of order $n^a$ when $n$ is large''%
\footnote{\label{foot:vecbigO}
    More rigorously, we mean that $\|v\|^2 / n = \Theta(n^{2a})$.
    Note this is \emph{different} from the common interpretation that $\|v\| = \Theta(n^{a})$.
};
likewise for $o(n^a)$, etc.
Recall the notations $x_t = x_t(\xi_t), \tilde x_t = x_t(\tilde \xi), \delta x_t = \sqrt n(x_t - x_{t-1})$ and likewise for $h_t, \tilde h_t, \delta h_t$.
The key insights are as follows:
\paragraph{Preliminary Calculations}
It turns out $\tilde x_{t+1} - \tilde x_t = \Theta(1/\sqrt n) = o(1)$ so $\delta \tilde x_{t+1} = \sqrt n (\tilde x_{t+1} - \tilde x_t))$ has $\Theta(1)$ coordinates.
Likewise for $\delta \tilde h_{t+1}$.
Consequently, for any $t$ and input $\xi$,
by telescoping,
\begin{align}
    h_t(\xi)
        &= h_0(\xi) + o(1).
        \label{eqn:g0}
\end{align}
Using $\nabla_u f = \frac{1}{\sqrt n} dh_t \xi_t^\top$
and $dh_t = \phi'(h_t) \odot v$, we have:
\begin{align}\label{1delta}
    \delta\tilde{h}_{t+1}
    = -\chi_t\xi_t^\top \tilde{\xi} \phi'(h_t)\odot v.
\end{align}
Also, $\delta\tilde{x}_{t+1} = \sqrt{n}\big(\phi(\tilde{h}_t + \frac{\delta\tilde{h}_{t+1}}{\sqrt{n}}) - \phi(\tilde{h}_t)\big)$. 
Since $\delta \tilde h_{t+1} = \Theta(1)$, by intuitive Taylor expansion, we have     
\begin{align}\label{1limit}
    \delta\tilde{x}_{t+1} &\approx \phi'(\tilde{h}_t)\odot \delta\tilde{h}_{t+1}.
\end{align}
The change in the output on example $\tilde{\xi}$ from step $t$ to step $t+1$ is given by:
\begin{align}\label{1dynamics}
    \tilde{f}_{t+1} - \tilde{f}_t = V^\top (\tilde{x}_{t+1} - \tilde{x}_t) = v^\top \delta\tilde{x}_{t+1}/n
\end{align}

\paragraph{IID Coordinates}
By definition $v$ has iid coordinates. It turns out $h_t(\xi), \delta h_t(\xi)$ (likewise for $x$) all have approx.\ iid coordinates of size $\Theta(1)$ as well.%
\footnote{Technically, they have iid coordinates only after conditioning on the initial function (GP) $f_0$.
Likewise, when we take expectation in this example (e.g.\ \cref{eqn:1LPexpectation,{eqn:dot}}), it's a conditional expectation of this kind.
See \cref{subsub:first_pass} for a rigorous treatment.
However, to convey the main intuition, we gloss over this technicality here.
\label{foot:condition}}
Let $Z^{\delta \tilde x_t}, Z^v$ denote the random variables encoding the corresponding coordinate distributions; likewise for the other vectors.
Note that $Z^{\delta \tilde x_t}, Z^v$ will in general be correlated, reflecting the coordinatewise correlation between $v$ and $\delta \tilde x_{t}$.

\paragraph{Law of Large Numbers}
and \cref{1dynamics,1delta} imply,
\begin{align}
    &\phantomeq\lim_{n\to\infty} \tilde{f}_{t+1} - \tilde{f}_t
    = \EV Z^v Z^{\delta \tilde x_{t+1}}
    \label{eqn:fdelta=EZZ}
    \\
    &= -\mathring\chi_t \xi_t^\top \tilde{\xi} \EV \phi'(Z^{\tilde h_t}) \phi'(Z^{h_t}) (Z^v)^2.
    \label{eqn:1LPexpectation}
\end{align}
where $\mathring\chi_t = \mathcal L_t'(\lim_{n} f_t)$ as in \cref{claim:NTK}.

\paragraph{Kernel Expression as Gaussian Expectation}
By \cref{eqn:g0}, in the $n\to\infty$ limit, $Z^{\tilde h_t} = Z^{h_0(\tilde \xi)}$ and $Z^{h_t} = Z^{h_0(\xi_t)}$.
They are independent from $Z^v$ and jointly Gaussian with variances $\|\tilde \xi\|^2, \|\xi_t\|^2$ and covariance $\tilde \xi^\trsp \xi_t$.
So (using the initialization of $v$ to simplify $\EV (Z^v)^2 = 1$),
\begin{align}
    \lim_{n\to\infty} \tilde{f}_{t+1} - \tilde{f}_t
    &= -\mathring\chi_t \xi_t^\top \tilde{\xi} \EV \phi'(Z^{h_t}) \phi'(Z^{\tilde{h}})\label{ntk1}
\end{align}
This can easily be seen to be \cref{kernel_d_1} (recall we assumed for simplicity the output layer $V$ is not trained).

\paragraph{Summary}
Our strategy so far has been computing the form of $Z^{\delta \tilde{x}_{t}}$ and plugging it into \cref{eqn:fdelta=EZZ} to compute the dynamics of the output in the limit.
Note that our approach differs from previous work which mainly focus on proving a bound on the change of the NTK post training.
As the architecture gets more complex, bounding the NTK movement becomes quite complex, but our approach easily scales due to the automation provided by the \emph{Tensor Programs} framework (see \cref{sec:TP}).

In the previous example, the coordinate distribution $Z^{\tilde{x}_{t}}$ took a fairly simple form, which allowed us to intuitively compute the expectation $\EV Z^{\tilde{x}_{t}} Z^v$. Before introducing a method for computing coordinate distributions in a general architecture, we move on to a slightly more involved architecture, with the intention of highlighting the intuition behind the general case. 

\subsection{2 Hidden Layers}
\label{sec:2LP}

In this example we consider a model of the form:
\begin{gather*}
f = V^\top x ,~~~x = \phi(h) ,~~~ h = W z\\
z = \phi(g) ,~~~ g = U\xi
\end{gather*}
where $U = \frac{u}{\sqrt{d}} \in \mathbb{R}^{n \times d},~W = \frac{w}{\sqrt{n}} \in \mathbb{R}^{n \times n},~V = \frac{v}{\sqrt{n}} \in \mathbb{R}^{n \times 1}$, for trainable parameters $u,w$, initialized iid from a normal distribution. As before we assume the last layer is not trained, and $d=1$.

Again, we want to establish \cref{claim:NTK} for this model.
As in the 1-hidden-layer example, the dynamics of the output in the limit is still given by \cref{eqn:fdelta=EZZ}.
This time around, the second hidden layer adds nontrivial complexity when evaluating the expectation $\EV Z^vZ^{\delta \tilde{x}_{t+1}}$. As we shall see, this complexity arises from the dependency of $\tilde{x}$ on the $n\times n$ matrices $w$ and $w^\top$, which will make it \emph{wrong} to naively apply LLN arguments. Resolving this complexity will pave the way to a general strategy which we will then be able to apply in any arbitrary architecture.
We now apply the same insights as in the 1-hidden-layer MLP. Namely:

\begin{itemize}
    \item \cref{eqn:g0} continues to hold with $h$ replaced by any of $\{x,h,z,g\}$.
\item After some brief calculations, with $dh_t$ denoting the scaled gradient $\sqrt n \nabla_{h_t} f$,
    \begin{align}
        \delta\tilde{g}_{t+1}
            &\approx -\chi_t\xi_t^\top \tilde{\xi}  \phi'(g_t) \odot \big(W^\top dh_t\big)\label{deltap}\\
        \delta\tilde{h}_{t+1}
            &\approx W\delta \tilde{z}_{t+1} - \chi_t \frac{z_t^\top \tilde{z}_t}{n}\phi'(h_t)\odot v \label{deltag}.
    \end{align}
    
    \item 
    As in the 1-hidden-layer case, for all $x \in \{g,z,h,x\}$, $x_t(\xi),\delta x_t(\xi)$ have iid coordinates of size $\Theta(1)$, as does $v$ by definition.%
    \footnote{Again, this is technically true only after conditioning on $f_0$; see \cref{foot:condition}.}
    Let $Z^{\delta \tilde x_t}, Z^v$ denote the (generally correlated) random variables encoding the corresponding coordinate distributions.
    \item As in \cref{1dynamics}, by naive Taylor expansion we have:
    \begin{align}
        \delta\tilde{z}_{t+1} &\approx \phi'(\tilde{g}_{t})\odot \delta\tilde{g}_{t+1}.\label{deltaq}
    \end{align}
    \item \cref{1dynamics,1limit} in the 1-hidden-layer case continue to hold here.
    Then by \cref{deltag} and Law of Large Numbers,
    \begin{align}
        &\phantomeq\lim_{n\to\infty} \tilde{f}_{t+1} - \tilde{f}_t
        = \EV Z^v Z^{\delta \tilde x_{t+1}}
        \label{eqn:fdelta=EZZ_2}
        \\
        &= - \mathring \chi_t \xi_t^\top \tilde{\xi}
        \EV [Z^{z_t}Z^{\tilde{z}}]
        \EV [\phi'(Z^{h_t})\phi'(Z^{\tilde{h}})]
        \label{eqn:2LPlayer2NTK}\\
        &\phantomeq- \mathring \chi_t \xi_t^\top \tilde{\xi}\EV[\phi'(Z^{\tilde{h}_t})Z^{W\delta \tilde{z}_{t+1}}Z^v].
        \label{eqn:1LPexpectation_2}
    \end{align}
    where $\mathring\chi_t = \mathcal L_t'(\lim_{n} f_t)$ as in \cref{claim:NTK}.
\end{itemize}
In this expression, the first term (\cref{eqn:2LPlayer2NTK}) can easily be seen to correspond to the contribution from $w$ to the NTK.
It remains to show that the second (\cref{eqn:1LPexpectation_2}) corresponds to the contribution from $u$.

\paragraph{Challenge of Analyzing \cref{eqn:1LPexpectation_2}}
To do this, we must reason about the coordinate distribution of $W\delta\tilde{z}_{t+1}$ (encoded by random variable $Z^{W\delta\tilde{z}_{t+1}}$) and compute the expectation in \cref{eqn:1LPexpectation_2}.
To understand why this represents a greater challenge than it might first appear, note that from $\delta\tilde{z}_{t+1} \approx \phi'(\tilde{g}_{t})\odot \delta\tilde{g}_{t+1}$ (\cref{deltaq}), the term ${W\delta\tilde{z}_{t+1}}$ hides within itself a dependency on $W^{\top}dh_t$ through $\delta\tilde{g}$ (\cref{deltap}).
While at $t=0$, we may assume $W^\trsp$ be independent of $W$ and obtain the correct results (Gradient Independent Assumption \citep{yangMFResnet,yang2}), this is no longer the case for $t>0$:
$Z^{W\delta\tilde{z}_{t+1}}$ will be nontrivially correlated with $\phi'(Z^{\tilde{h}_t})$ and $Z^v$ (which would be false if $W^\trsp$ can be assumed independent of $W$).
We will give some intuition why later in \cref{eqn:WWtrspExample}.
Now, what is this dependence exactly?
\begin{restatable}{claim}{claimZdot}
    \label{claim:Zdot}
    Based on the above discussion and some easy calculations, $\delta\tilde{z}_{t+1}$ can be written as $\Phi(W^\top dh_t)$ for some $\Phi: \R \to \R$ applied coordinatewise (which will depend on other vectors not of the form $W^\trsp \bullet$).
    Then it turns out\footnotemark[\value{footnote}]
    \begin{align}\label{eqn:dot}
        Z^{W\delta\tilde{z}_{t+1}} = G + Z^{dh_t}\EV \frac{\partial Z^{\delta\tilde{z}_{t+1}}}{\partial Z^{W^\top dh_t}},
    \end{align}
    where $G$ is some Gaussian variable independent from $Z^v$, and $\frac{\partial Z^{\delta\tilde{z}_{t+1}}}{\partial Z^{W^\top dh_t}} \defeq \Phi'(Z^{W^\top dh_t})$.
\end{restatable}
\paragraph{Getting \cref{claim:NTK}}
Thus, from
\cref{deltaq,deltap}, it follows that:
\begin{align}
Z^{\delta\tilde{z}_{t+1}} &= -\mathring \chi_t \xi_t^\top \tilde{\xi}\phi'(Z^{\tilde{g}_t})\phi'(Z^{g_t})Z^{W^\top dh_t} \label{eqn:delta_z}\\
\EV\frac{\partial Z^{\delta\tilde{z}_{t+1}}}{\partial Z^{W^\top dh_t}} &= -\mathring \chi_t \xi_t^\top \tilde{\xi}\EV[\phi'(Z^{\tilde{g}_t})\phi'(Z^{g_t})]\label{eqn:delta_z_dot_exp}.
\end{align}
Plugging into \cref{eqn:fdelta=EZZ_2,{eqn:dot}}, followed by some straightforward calculation, then yields \cref{claim:NTK}.

\newcommand{\xx}{\mathbf{x}}
\newcommand{\yy}{\mathbf{y}}
\newcommand{\zz}{\mathbf{z}}
\paragraph{Intuition behind \cref{claim:Zdot}}
\cref{eqn:dot} may appear cryptic at first, so let's give some intuition using an example.
Suppose $\Phi$ in \cref{claim:Zdot} is actually identity.
For brevity, we set $\xx = dh_t, \yy = \delta \tilde z_{t+1}  \in \R^n$.
Then, following straightforward calculation, $W \yy = W W^\trsp \xx$ has coordinates
\begin{align}
    (W\yy)_\alpha
    = 
        \sum_{\gamma \ne \alpha} \xx_\gamma \sum_{\beta=1}^n W_{\alpha\beta} W_{\gamma\beta} 
        + \sum_{\beta=1}^n (W_{\alpha\beta})^2 \xx_\alpha
        \label{eqn:WWtrspExample}
\end{align}
Now, the second sum converges via LLN to $\xx_\alpha$ as $n\to\infty$.
On the other hand, the first sum will converge via CLT to $\Gaus(0, \lim \|\xx\|^2/n)$.
Thus, in terms of $Z$s, we have
\begin{align}
    Z^{W\yy} = G + Z^\xx = G + Z^\xx \EV \Phi'
\end{align}
for some Gaussian $G$; this corresponds directly to \cref{eqn:dot}.%
\footnote{
    This example was worked out in \citep{yang2,yang3} as well, though in different contexts.
    Readers needing more explanation may see those works.
}
For general $\Phi$, a similar intuition applies after Taylor expansion of $\Phi$.

\paragraph{Summary}
This 2-hidden-layer example proceeded much the same as the 1-hidden-layer case, with the main exception of analyzing the interaction of the $n\times n$ Gaussian matrix $W$ and $W^\trsp$ (\cref{eqn:1LPexpectation_2}) that occurs after taking at least 1 step of SGD.
This was absent in the 1-hidden-layer case because each weight matrix has at most one side tending to infinity.
Such analysis is crucial to obtaining the right results, as assuming $W^\trsp$ be independent from $W$ would imply $f$ does not move from initialization.%
\footnote{One can see this easily by modifying our calculations above.}

It turns out these two examples have essentially covered all of the core ideas needed to extend the analysis into arbitrary architectures.
To formalize and scale up our calculations, we now turn to the \emph{Tensor Programs} framework.

\begin{figure}[t]
\centering
\includegraphics[width=0.45\textwidth]{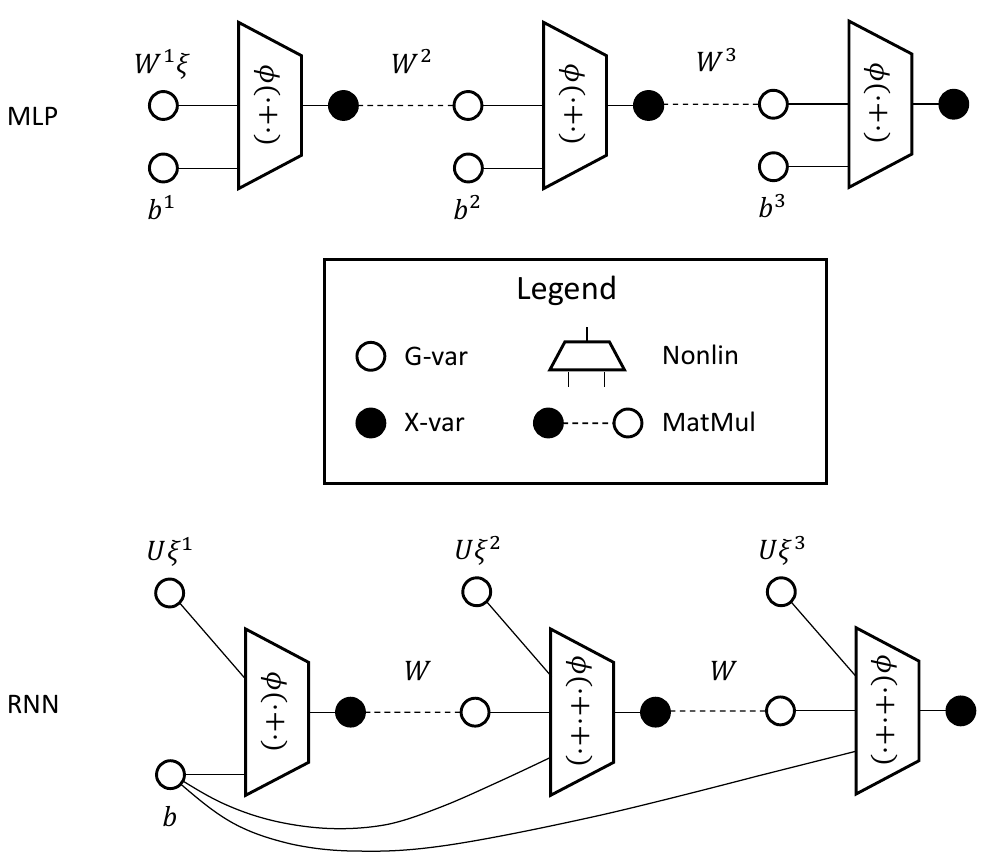}
\caption{\textbf{Graphical form of \netsort{} programs for MLP and RNN.}
An empty node corresponds to a G-var (an initial vector or a vector created by \refMatmul{}), while a filled node corresponds to an X-var (a vector created by \refNonlin{}).
Each dashed edge corresponds to matrix multiplication by the labeled matrix.
Each gate (connected by solid edges) represents a \refNonlin{} application.
The computation follows the direction of the gates (left-to-right here).
Note we have $W^1 \xi$ instead of $\xi$ as an initial vector because we only allow $\R^n$ vectors; likewise for $U \xi^i$.
For the same reason, we don't express the network output in this graph.}
\label{fig:reduced_programs}
\end{figure}

\section{Tensor Programs}\label{sec:TP}

So far, our results have been obtained by unrolling the SGD updates on toy models with specific architectures, and using informal arguments. Obviously, these computations quickly become unmanageable when the architecture becomes more complex.
The sheer amount of architectural innovations that have sprung up in recent years requires us to adopt a much more general formulation of our results.
To that end, we adopt the Tensor Programs (TP) framework developed in \cite{yang,yang2,yang3}.
In a nutshell, it provides a language for describing typical computations done in the context of neural networks, such as forward and backward propagation.
It is simultaneously simple and expressive, covering all standard architectures \citep{yang,yang2}.
Here we review two basic forms of Tensor Programs, \netsort{} and \netsortp{}.
\enlargethispage{\baselineskip}
\begin{defn}\label{defn:simpleNetsorT}
    A \netsort{} program is just a sequence of $\R^n$ vectors inductively generated via one of the following instructions from an initial set $\mathcal{V}$ of random $\R^n$ vectors and a set $\mathcal{W}$ of random $n\times n$ matrices%
    \begin{description}
        \item [\normalfont\textsc{Nonlin}\label{instr:nonlin}] For $x^{1},\ldots,x^{k}\in\R^{n}$ in the program and any $\psi:\R^{k}\to\R$, we can generate $\psi(x^{1},\ldots,x^{k})\in\R^{n}$
        \item [\normalfont\textsc{MatMul}\label{instr:matmul}] Given $W\in\R^{n\times n}$ and $x\in\R^{n}$, we can generate $Wx\in\R^{n}$ or $W^{\trsp}x\in\R^{n}$
    \end{description}
\end{defn}

\paragraph{Graphical Form}
We propose to represent a \netsort{} program as a computational graph, where each node in the graph represents vectors (initial or generated), each (dashed) edge represents a \refMatmul{}, and each gate represents a \refNonlin{}.
For example, \cref{fig:reduced_programs} shows the computation graphs expressing (the forward passes of) an MLP and an RNN.
We can also express the backpropagation as well (see \cref{fig:backprop}).
Graphically, the initial vectors are the empty nodes with only one edge coming out, toward the direction of computation.
The matrices correspond to (the labels of) the dashed edges.
We can also define the \emph{output vectors} to correspond to the nodes that have only one edge coming out, \emph{against} the direction of computation.

\paragraph{Neural Network Representation}
Each \netsort{} program can be thought of as computing a function $(\R^n)^{\mathcal V} \times (\R^{n\times n})^{\mathcal W} \to \R^{\mathcal Y}$ taking an instantiation of the initial vectors $\mathcal V$ and matrices $\mathcal W$ and computing the values of all output vectors $\mathcal Y$.
We can say a program \emph{represents a network} if it computes the \emph{body} of it (without the input and output layers), as exemplified by \cref{fig:reduced_programs}.
This is formalized below.
\begin{defn}\label{defn:nnrep}%

    Consider a neural network $f: (\R^d)^k \to \R$ with input embedding matrices $U^1, \ldots, U^k \in \R^{n\times d}$ (not necessarily distinct) and readout matrix $V \in \R^n$, so that $f(\xi^1, \ldots, \xi^k) = V^\trsp \Phi(U^1 \xi^1, \ldots, U^k \xi^k; \Theta)$ for some function $\Phi(x^1, \ldots, x^k; \Theta)$ with parameters $\Theta$.
    We say a \netsort{} program \emph{represents} $f$ if it computes $\Phi$ (under some correspondence of $\mathcal V \cup \mathcal W$ to $\{x^1, \ldots, x^k\}\cup \Theta$).%
    \footnote{
        We only consider $f$ with scalar output for simplicity but generalization to multi-dimensional output is straightforward.
    }
    
\end{defn}
For example, the programs in \cref{fig:reduced_programs} resp.\ represent a 3-hidden-layer MLP and an RNN running for 3 steps.
Note that the initial vectors correspond to a combination of input embeddings (e.g.\ $W^1 \xi$) and vector parameters (e.g. biases) and the matrices correspond to matrix parameters (e.g.\ weights).

\paragraph{Intuition for a Program in the Large $n$ Limit}
Typically, the vectors (resp.\ matrices) in a program will be sampled iid like $\Gaus(0, 1)$ (resp.\ $\Gaus(0, 1/n)$), corresponding to the ``standard'' initialization of neural networks.%
\footnote{
    In the original definition of \citep{yang,yang2,yang3}, the vectors can have correlations between them, but we can always rewrite these vectors as linear image of another set of uncorrelated vectors.
}
In such cases, when $n\to\infty$, a program behaves as follows, in a gist:
\begin{description}
    \item[IID Coordinates] Any vector $x \in \R^n$ in the program has roughly iid coordinates.
    We write $Z^x$ for the random variable encoding this coordinate distribution.
    This $Z^x$ may be correlated with $Z^y$ for other vector $y$ in the program, such that, for example, $\lim_{n\to\infty} x^\trsp y/n = \EV Z^x Z^y$.
    \item[Nonlin] $Z^{\psi(x^1, \ldots, x^k)} = \psi(Z^{x^1}, \ldots, Z^{x^k})$.
    \item[MatMul, without $(W, W^\trsp)$-Interaction] Consider a matrix $W \in \mathbb{R}^{n \times n}$ 
     in the program and any set of $\mathbb{R}^n$ vectors $\X$ not dependent on vectors of the form $W^\trsp \bullet$.
    Then the set of random variables $\{Z^{Wx}: x \in \X\}$ are jointly Gaussian with mean zero and covariance $\Cov(Z^{Wx}, Z^{W\tilde{x}}) = \EV Z^x Z^{\tilde{x}}$ for any $x,\tilde{x} \in \X$.
    If $\bar W \ne W$ is another matrix in the program and $\Y$ is a set of such vectors w.r.t.\ $\bar W$, then the set $\{Z^{Wx}: x \in \X\}$ is independent from $\{Z^{\tilde{W}y}: y \in \Y\}$.
    \item[MatMul, with  $(W, W^\trsp)$-Interaction]
    For general $x$, $Z^{Wx}$ decomposes into a sum of a Gaussian part, identical to $Z^{Wx}$ in the above case, and a correction term.
    This decomposition is a generalization of \cref{eqn:dot}.
\end{description}

See \cref{thm:PLNetsorT+MasterTheorem} for formal details.

\paragraph{\netsortp{} Programs}
\citep{yang,yang2} showed that \netsort{} suffices to express the forward and backward passes of most architectures such as ResNet (with Batchnorm), but Transformer and other standard architectures require adding to \netsort{} a new ``averaging'' instruction%
\footnote{For readers familiar with Tensor Programs, this is equivalent in expressivity to \textsc{Moment}, which is a composition of a \refNonlin{} and this averaging instruction.}
that returns the ``empirical average'' $\f 1 n \sum_{\alpha=1}^n x_\alpha$ of a vector $x \in \R^n$.
In the $n\to\infty$ limit, this scalar converges to $\EV Z^x$ as would be expected from the intuitions above.
This extension of \netsort{} (called \netsortp{}) also allows us to express the network output and loss derivative (e.g.\ in contrast to \cref{fig:reduced_programs}), which will be a technical requirement for unrolling the entirety of SGD training inside a single program, a key step in the proof of our main result.
See discussion of proof formalization in \cref{sec:ProvingNTKtrain}.
We can say a \netsortp{} program \emph{represents a network $f$} if it computes the body of $f$.

\section{Universality of Kernel Dynamics}\label{sec:ProvingNTKtrain}

\citep{yang,yang2} showed that any neural network of standard architecture is represented by a \netsortp{} program.
Moreover, 
\begin{thm}[\citet{yang2}]
    For a neural network as in \cref{setup:main} below, its Neural Tangent Kernel at initialization has a well-defined infinite-width limit $\mathring{\mathcal K}$.
\end{thm}

\begin{setup}[Representable NN in NTK Parametrization]\label{setup:main}
    Suppose a neural network $f$ is represented by a \netsortp{} program (in the sense of \cref{defn:nnrep}) whose \refNonlin{} all have polynomially bounded derivatives.%
    \footnote{More generally, we can allow any pseudo-Lipschitz function here, but for simplicity we go with the statement in the main text.}
    Adopt the NTK parametrization: for every matrix parameter $W \in \R^{n\times n}$ of $f$, we factor $W = \f 1 {\sqrt n} w$ where $w$ is the trainable parameter;
    likewise, for each input layer matrix $U^i \in \R^{n\times d}$, we factor $U^i = \f 1 {\sqrt d} u^i$,
    and likewise the output matrix $V = \f 1 {\sqrt n} v$, such that $u^i, v$ are trainable.
    Finally, we randomly initialize all trainable parameters iid as $\Gaus(0, 1)$.
\end{setup}

Our main result is to show that the SGD training of such a neural network described in \cref{setup:main} reduces to kernel gradient descent with kernel $\mathring{\mathcal K}$ in the infinite-width limit.

\newcommand{\loss}{\mathcal{L}}
\begin{thm}[\ref{NTKtrain} is Architecturally Universal]\label{thm:main}
    Consider training a network $f$ described in \cref{setup:main} via SGD with batch-size 1 and (WLOG) learning rate 1.
    Let $\xi_t$ be the input and $\loss_t: \R \to \R$ be the loss function (absorbing the label) at time $t$.
    Suppose $\loss_t$ is continuous for all $t$.
    Then, for any $\xi$ and $t$, $f_t(\xi)$ converges almost surely to a random variable $\mathring f_t(\xi)$ as width $\to \infty$, such that
    \begin{align}\label{eq:ntkdynamics}
        \mathring f_{t+1}(\xi) - \mathring f_t(\xi) = - \mathring{\mathcal K}(\xi, \xi_t) \loss_t'(\mathring f_t(\xi_t))
    \end{align}
    where $\mathring{\mathcal{K}}$ is the infinite-width NTK (at initialization) of the neural network.
    \end{thm}

The full proof of \cref{thm:main} is given \cref{sec:appendix_main_proof}.

\paragraph{Extension}
We briefly mention several ways our result can be easily extended.
\emph{0) Different batch sizes, learning rate schedules, and nonscalar outputs}.
\emph{1) Variants of NTK parametrization.}
We can deal with any parametrization that scales the same way as NTK parametrization, e.g.\ weights are sampled like $N(0, \sigma^2)$ for any $\sigma$, with the multipliers $\gamma/fanin$ for any $\gamma$.
\emph{2) Variable width}.
In real networks, the width of different layers can often be different (e.g. in ResNet).
Our result can be extended to the case where the widths tend to infinity at a fixed ratio, using the variable-width version of Tensor Programs \citep{yang3}.
\emph{3) Unsupervised and other learning settings}
can be covered because their training and testing computation can be written into Tensor Programs.
\emph{4) Weight decay, momentum, and other optimizer tricks}
can be covered as well as they can be straightforwardly written into Tensor Programs, but in general the kernel will change from step to step in contrast to \cref{thm:main}.

\subsection{Proof Sketch of Special Case}
\label{sec:proofsketch}
To convey the main idea, we give a proof sketch of a simplified problem: we assume
\begin{enumerate*}[label=\arabic*)]
    \item the input, output layers and biases are not trained (the network has only $\R^{n\times n}$ matrices as trainable parameters);
    \item the forward pass does not contain both a weight matrix $W$ and its transpose $W^\trsp$ (but a single matrix $W$ can still be used multiple times without being transposed);%
    \item input space is $\R^d$ (with $k=1$), and $f = V^\top x \in \R$;
    \item the output vector $x$ is a G-var;
    \item the network is represented by a \netsort{} (instead of \netsortp{}) program;
    \footnote{This is not common in deep learning, but appears in some weight-tied autoencoders \citep{tied-encoder}.}
\end{enumerate*}
In the appendix, we prove the general case with these simplifications lifted.

It turns out, every \netsort{} can be simplified into a \emph{standard form} of sorts, which greatly facilitates our proof.
\begin{defn}
    In a \netsort{} program, a \emph{G-var}%
    \footnote{``G'' because G-vars often are roughly Gaussian vectors}
    is an initial vector or a vector created by \refMatmul{}, while an \emph{X-var} is a vector created by \refNonlin{}.%
    \footnote{\emph{Var} is short for \emph{variable}, as the vectors are considered \emph{variables} in the program.
        In previous works, \emph{H-var} refers to any vector in the program; we will not use this terminology.}
    We define a \emph{reduced \netsort{} program} as a program in which only G-vars are allowed as inputs to a \refNonlin{}, while only an X-var is allowed as input to a \refMatmul{}.
\end{defn}
Observe that any \netsort{} program may be trivially expressed as a reduced \netsort{} program by: 1) collapsing chains of non-linearities which appear consecutively, and
2) insert a \refNonlin{}  operation with $\psi(x) = x$ in between consecutive G-vars.
Hence, we may safely assume that $f$ is representable by a reduced \netsort{} program.

\paragraph{Key Idea: Paths}
The examples of \cref{sec:1LP,sec:2LP} exposed several insights, such as the iid-coordinates intuition, important for proving \cref{thm:main}.
Now we discuss the one remaining key idea for scaling up to general architectures:

\begin{figure}[t]
    \centering
    \includegraphics[width=0.45\textwidth]{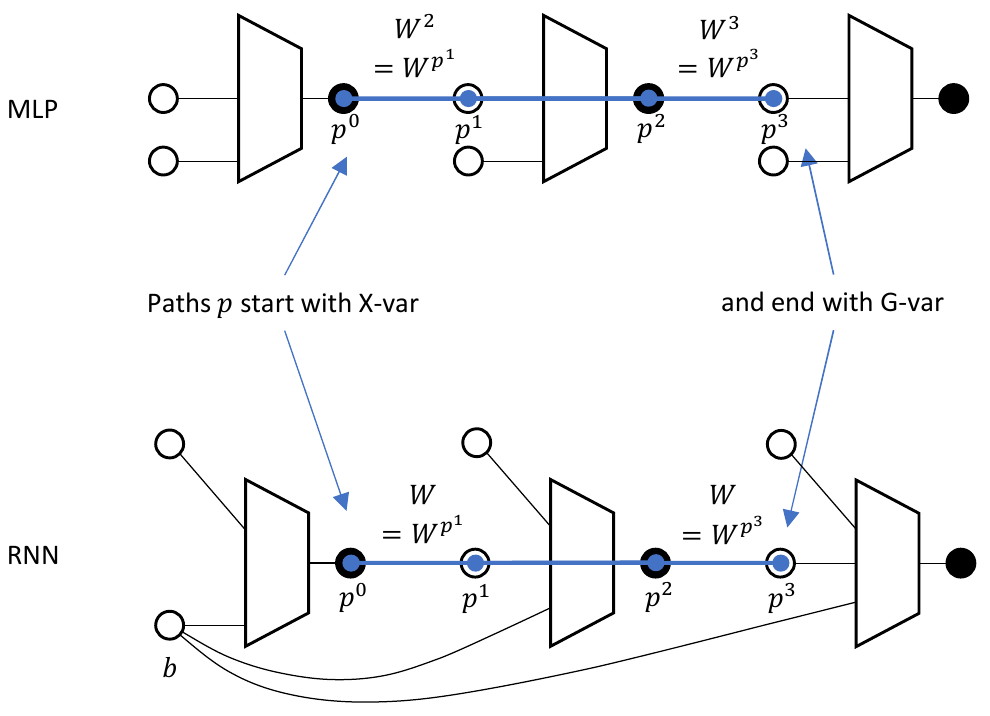}
    \caption{\textbf{Paths.}
    A graphical illustration of paths in an MLP and RNN. A path $p$ always starts with an X-var and ends with a G-var.
    In these examples, path length $|p| = 4$.}
    \label{fig:paths}
\end{figure}

\begin{figure}[t]
    \centering
    \includegraphics[width=0.45\textwidth]{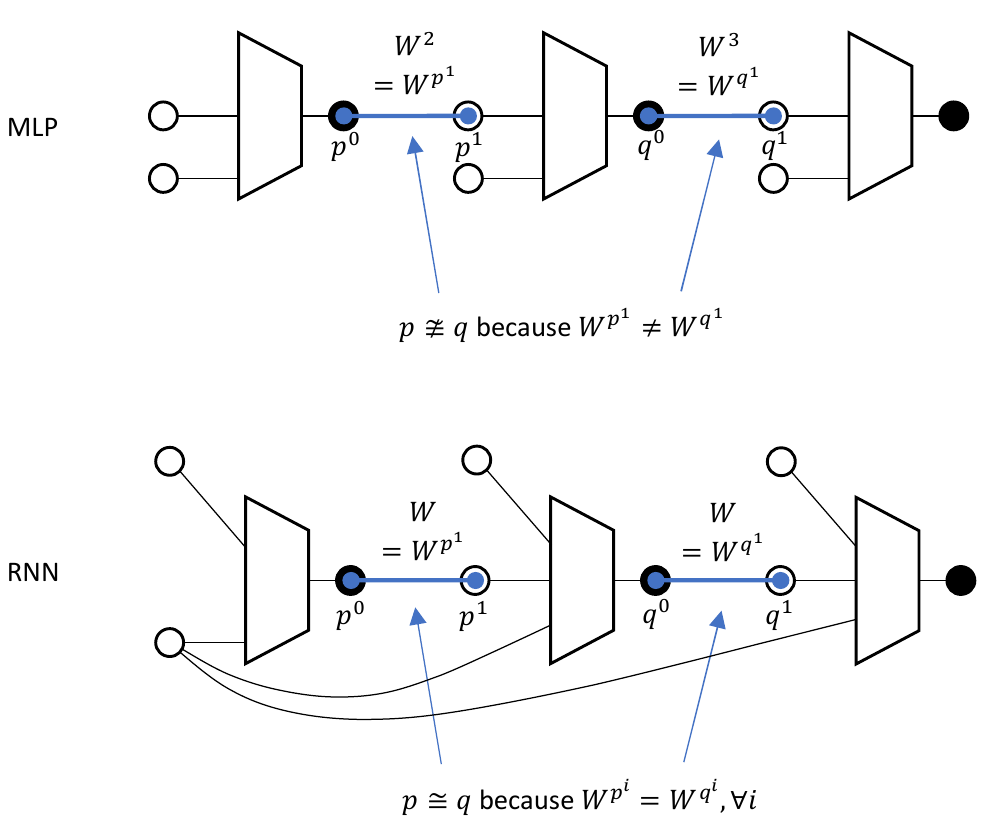}
    \caption{\textbf{Paths Isomorphism.}
    A graphical illustration of path isomorphism for paths $p,q$ with $|p|=|q|=2$ in an MLP and RNN. In the MLP example, $p$ is not isomorphic to $q$ because the weights $W^1,W^2$ defining the vectors $p^1,q^1$ are not identical.
    For the RNN, $p$ is isomorphic to path $q$ because $p^1$ and $q^1$ are defined with the same weight $W$.}
    \label{fig:paths_iso}
\end{figure}

\begin{restatable}[Paths]{defn}{Paths}\label{defn:paths}
In a \netsort{} program, a path $p$ starts with an X-var and ends with a G-var, alternating between X- and G-vars along the path.
We write $p^{0}$ for the starting X-var, $p^{1}$ for the following G-var, and so on, as well as $p^{-1}$ for the ending G-var (see \cref{fig:paths} for a graphical illustration).
For odd $i$, let $W^{p^{i}}$ denote the defining matrix of G-var $p^{i}$. For two equal length paths $p,q$, we write $p\cong q$ (path $p$ is isomorphic to path $q$) if for all odd $i$, $W^{p^{i}}$ is the same matrix as $W^{q^{i}}$.%
\footnote{Here we are talking about equality of symbols rather than equality of values of those symbols.}
In other words, we say path $p$ is isomorphic to path $q$ if their sequences of \refMatmul{} matrices are identical, (but the \refNonlin{} don't have to be, see \cref{fig:paths_iso} for a graphical illustration). Let $|p|$ denote the number of vectors in $p$ (this is always an even number).
\end{restatable}

The collection of paths $p$ starting with an X-var $p^0 = x$ and ending with a G-var $h$ describes all possible pathways of backpropagating an error signal $dh$ at $h$ to an error signal $dx$ at $x$.
Simultaneously, it also describes all possible pathways of forward propagating a change in $x$ to a change in $h$.

\paragraph{Decomposing the Change in $f$ to a Sum over Paths}
Because the gradient $\nabla_W f$ of a weight $W \in \R^{n\times n}$ is the sum of outer products $\sum_{h,x: h=Wx} dh\otimes x$, summing over all G-vars $h$ and X-vars $x$ in the program with $h= Wx$ (where $dh$ denotes $\nabla_h f$), we also have
\[\nabla_W f = \sum_{h,x: h=Wx} \sum_p J^p\otimes x.\]
where
\[(J^p)^\trsp = \frac{\partial f}{\partial p^{-1}}\times\frac{\partial p^{-1}}{\partial p^{-2}}\times\frac{\partial p^{-2}}{\partial p^{-3}}\times...\times \frac{\partial p^{2}}{\partial h}\]
i.e, $J^p$ denotes the error signal at $h$ from  backpropagation through path $p$, and $p$ ranges over all paths starting with $p^0 =x, p^1=h$ and ending with the output node of the underlying program.
Recall $W$ factors as $\f 1 {\sqrt n} w$ where $w$ is the trainable parameter, not $W$.
By the discussion above, updating $w$ with $\nabla_w f = \f 1 {\sqrt n} \nabla_W f$  causes $f$ to change by
\begin{equation}
    \begin{gathered}
            \left\langle \f 1 {\sqrt n} \nabla_W f, \f 1 {\sqrt n} \nabla_W f \right\rangle
            \\
        =
            \f 1 n \sum_{h,x: h=Wx} \sum_{\bar h,\bar x: \bar h=W\bar x} \sum_p \sum_{\bar p} \langle J^p, J^{\bar p} \rangle \langle x, \bar x \rangle.
            \label{eqn:JJ}
    \end{gathered}
\end{equation}
When every parameter $w$ is randomly initialized iid as $\Gaus(0, 1)$, it turns out that $\langle J^p, J_{\bar p} \rangle$ will go to 0 as $n \to \infty$ unless $p \cong \bar p$ (\cref{defn:paths}).
If one think of $J^p$ as a product of random Gaussian matrices (interleaved with other matrices), then this is akin to the fact that, for a mixed moment $M \defeq \EV \prod_{i = 1}^r Z_{\gamma(i)}, \gamma: [r] \to [k]$ of standard iid Gaussians $Z_1, \ldots, Z_k$, $M$ is nonzero iff every $Z_i$ appears an even number of times in the product.
This means we can replace the 4 nested sums in \cref{eqn:JJ} with the single sum $\sum_{p \cong \bar p}$ and rewrite $x = p^0, \bar x = \bar p ^0$.

We have suppressed dependence on input in \cref{eqn:JJ}.
Being more explicit about it and performing updates on all weights, we have
\begin{equation}
    \begin{gathered}
        f_{t+1}(\xi) - f_t(\xi) \\
        \approx -\loss_t'( f_t(\xi_t))\sum_{p \cong \bar p} \langle J^{p}(\xi), J^{\bar p}(\xi_t) \rangle \langle p^0(\xi), \bar p^0(\xi_t) \rangle.    
        \label{eqn:deltaf}
    \end{gathered}
\end{equation}
after taking a gradient step on input $\xi_t$ at initialization $t=0$.
(Here $p^0(\xi)$ denotes the vector $p^0$ as a function of $\xi$ at initialization).
However, \cref{eqn:JJ} holds for general $t$ as well:
The key insight is similar to \cref{eqn:g0} in the 1-hidden-layer example, that vectors $p^0(\xi)$, $J^{p}(\xi)$, etc change vanishingly from their initial values as $n \to \infty$, after any number of SGD steps.
Because our arguments above only depend on inner products between vectors, this means that the error in  \cref{eqn:deltaf} for $t > 0$ vanishes as $n \to \infty$.
Finally, at least heuristically, it is straightforward to show the RHS of \cref{eqn:deltaf} is the NTK via a series of calculations exemplified by those in \cref{sec:1LP,sec:2LP}, to get \cref{eq:ntkdynamics}.

\paragraph{Putting It All Together and Proof Formalization}
While the core ideas discussed above are intuitive, making them rigorous at face value would be quite challenging.
Instead we use the machinery offered by the Tensor Programs framework.
The mechanics of the proof then goes as follows:
1) First we unroll SGD of $f$ into a \netsortp{} program.%
\footnote{\label{footnote:relyonTP3}
    We note that this formalization crucially relies on \netsortp{} and its Master Theorem from \citep{yang3} because the SGD unrolling cannot be done in \netsort{}.
    The reason is that we need to express the output and loss derivatives of the network, which are scalars (or at least finite dimensional), and that cannot be done in a \netsort{} program.
    Furthermore, the Master Theorem from \citep{yang2} only pertains to a specific type of programs that look like the \emph{first} backpropagation after initialization.
    Thus, it cannot deal with the complete unrolling of SGD as we do here, which requires the more advanced Master Theorem from \citep{yang3}.
}
This is similar to the equations in \cref{sec:1LP,sec:2LP}; the key here is to express $\delta x_{t+1} = \sqrt n(x_{t+1} - x_t)$ as a vector in the program, for any (pre-)activation $x$.
2) We apply the \emph{\netsortp{} Master Theorem} \citep{yang3} to this program.
This yields the coordinate distribution of each vector.
The core insights here are demonstrated by the calculation with the $Z$ random variables in \cref{sec:1LP,sec:2LP}.
3) Finally, we need to show that (the rigorous version of) \cref{eqn:deltaf} indeed recovers the NTK and agrees with \cref{eq:ntkdynamics}.
This is done via an inductive (symbolic) computation, and the path concept in this section plays a key role here.

\section{Related Works}\label{sec:2}

The connection between kernel methods and neural networks has had a long history before its recent resurgence.
The Gaussian Process (NNGP) view of wide neural networks, which characterizes the behaviour of training only the last layer of a wide neural network, has been studied in \cite{gp1,gp2,gp3,gp4,gp5,gp6,gp7}. Since the original NTK paper \cite{NTK}, many works have informally derived the infinite-width NTK for various architectures such as CNNs \cite{arora}, RNN \cite{rnnntk}, attention \cite{attention_ntk}, ensembles \cite{ce} and graph neural networks \cite{graphntk}, but none of them formally proved \ref{NTKinit} or \ref{NTKtrain} for those architectures.
Finite width corrections to the NTK were derived for fully connected networks in \cite{boris,resntk}.
The validity of the NTK theory was empirically studied in \cite{wide} for a variety of architectures.

The Tensor Program framework \cite{yang,yang2,yang3} was introduced in an attempt to unify and generalize the NNGP/NTK theory to a broad range of architectures, eliminating the need to re-develop the theory for each new architecture.
For example, \cite{yang} proved the architectural universality of NNGP correspondence, while \cite{yang2} proved that of \ref{NTKinit}.
On the other hand, \cite{yang3} developed the most general machinery for Tensor Programs and as a corollary constructed a comprehensive theory of nonlinear random matrix theory, that, for example, can calculate the singular value distribution of a wide neural network of any architecture.
Our proofs depend on the machinery of \cite{yang3} crucially, as discussed in \cref{sec:ProvingNTKtrain}.

\enlargethispage{\baselineskip}

\section{Conclusion}

New theories of deep learning almost always start with MLPs, rightly so as they are the simplest case and can often reveal the key insights more clearly.
Of course, as deep learning itself is an applied field, one should always ask whether insights on MLPs extend to more general architectures, i.e.\ whether there is an \emph{architecturally universal extension} of a proposed theory of MLPs.
This is not always easy to answer.

In this paper, we showed that the NTK theory is architecturally universal, but more importantly, we showed that the Tensor Programs technique is a very powerful tool for answering the above question as a matter of routine.
Looking forward, we hope to apply it to generate more novel and general insights.

\newpage
\twocolumn
\bibliography{refs}

\begin{thebibliography}{43}
\providecommand{\natexlab}[1]{#1}
\providecommand{\url}[1]{\texttt{#1}}
\expandafter\ifx\csname urlstyle\endcsname\relax
  \providecommand{\doi}[1]{doi: #1}\else
  \providecommand{\doi}{doi: \begingroup \urlstyle{rm}\Url}\fi

\bibitem[Alemohammad et~al.(2020)Alemohammad, Wang, Balestriero, and
  Baraniuk]{rnnntk}
Alemohammad, S., Wang, Z., Balestriero, R., and Baraniuk, R.
\newblock The recurrent neural tangent kernel.
\newblock \emph{ArXiv}, abs/2006.10246, 2020.

\bibitem[Allen-Zhu et~al.(2018)Allen-Zhu, Li, and
  Song]{allen-zhu_convergence_2018}
Allen-Zhu, Z., Li, Y., and Song, Z.
\newblock A {Convergence} {Theory} for {Deep} {Learning} via
  {Over}-{Parameterization}.
\newblock \emph{arXiv:1811.03962 [cs, math, stat]}, November 2018.
\newblock URL \url{http://arxiv.org/abs/1811.03962}.

\bibitem[Arora et~al.(2019)Arora, Du, Hu, Li, Salakhutdinov, and Wang]{arora}
Arora, S., Du, S., Hu, W., Li, Z., Salakhutdinov, R., and Wang, R.
\newblock On exact computation with an infinitely wide neural net.
\newblock In \emph{NeurIPS}, 2019.

\bibitem[Bahdanau et~al.(2015)Bahdanau, Cho, and Bengio]{attention1}
Bahdanau, D., Cho, K., and Bengio, Y.
\newblock Neural machine translation by jointly learning to align and
  translate.
\newblock \emph{CoRR}, abs/1409.0473, 2015.

\bibitem[Bruna et~al.(2014)Bruna, Zaremba, Szlam, and LeCun]{graph1}
Bruna, J., Zaremba, W., Szlam, A., and LeCun, Y.
\newblock Spectral networks and locally connected networks on graphs.
\newblock \emph{CoRR}, abs/1312.6203, 2014.

\bibitem[Cho et~al.(2014)Cho, Merrienboer, Çaglar G{\"u}lçehre, Bahdanau,
  Bougares, Schwenk, and Bengio]{gru}
Cho, K., Merrienboer, B.~V., Çaglar G{\"u}lçehre, Bahdanau, D., Bougares, F.,
  Schwenk, H., and Bengio, Y.
\newblock Learning phrase representations using rnn encoder-decoder for
  statistical machine translation.
\newblock \emph{ArXiv}, abs/1406.1078, 2014.

\bibitem[Daniely et~al.(2016)Daniely, Frostig, and Singer]{gp1}
Daniely, A., Frostig, R., and Singer, Y.
\newblock Toward deeper understanding of neural networks: The power of
  initialization and a dual view on expressivity.
\newblock In \emph{NIPS}, 2016.

\bibitem[Defferrard et~al.(2016)Defferrard, Bresson, and Vandergheynst]{graph2}
Defferrard, M., Bresson, X., and Vandergheynst, P.
\newblock Convolutional neural networks on graphs with fast localized spectral
  filtering.
\newblock In \emph{NIPS}, 2016.

\bibitem[Du et~al.(2019)Du, Hou, P{\'o}czos, Salakhutdinov, Wang, and
  Xu]{graphntk}
Du, S., Hou, K., P{\'o}czos, B., Salakhutdinov, R., Wang, R., and Xu, K.
\newblock Graph neural tangent kernel: Fusing graph neural networks with graph
  kernels.
\newblock In \emph{NeurIPS}, 2019.

\bibitem[Du et~al.(2018)Du, Zhai, Poczos, and Singh]{du_gradient_2018}
Du, S.~S., Zhai, X., Poczos, B., and Singh, A.
\newblock Gradient {Descent} {Provably} {Optimizes} {Over}-parameterized
  {Neural} {Networks}.
\newblock \emph{arXiv:1810.02054 [cs, math, stat]}, October 2018.
\newblock URL \url{http://arxiv.org/abs/1810.02054}.

\bibitem[Duvenaud et~al.(2015)Duvenaud, Maclaurin, Aguilera-Iparraguirre,
  G{\'o}mez-Bombarelli, Hirzel, Aspuru-Guzik, and Adams]{graph3}
Duvenaud, D., Maclaurin, D., Aguilera-Iparraguirre, J., G{\'o}mez-Bombarelli,
  R., Hirzel, T., Aspuru-Guzik, A., and Adams, R.
\newblock Convolutional networks on graphs for learning molecular fingerprints.
\newblock In \emph{NIPS}, 2015.

\bibitem[Fukushima(1975)]{conv1}
Fukushima, K.
\newblock Cognitron: A self-organizing multilayered neural network.
\newblock \emph{Biological Cybernetics}, 20\penalty0 (3):\penalty0 121--136,
  1975.
\newblock \doi{10.1007/BF00342633}.
\newblock URL \url{https://doi.org/10.1007/BF00342633}.

\bibitem[Fukushima(1980)]{conv2}
Fukushima, K.
\newblock Neocognitron: A self-organizing neural network model for a mechanism
  of pattern recognition unaffected by shift in position.
\newblock \emph{Biological Cybernetics}, 36\penalty0 (4):\penalty0 193--202,
  1980.
\newblock \doi{10.1007/BF00344251}.
\newblock URL \url{https://doi.org/10.1007/BF00344251}.

\bibitem[Hanin \& Nica(2019)Hanin and Nica]{boris}
Hanin, B. and Nica, M.
\newblock Finite depth and width corrections to the neural tangent kernel.
\newblock \emph{ArXiv}, abs/1909.05989, 2019.

\bibitem[Hazan \& Jaakkola(2015)Hazan and Jaakkola]{gp2}
Hazan, T. and Jaakkola, T.
\newblock Steps toward deep kernel methods from infinite neural networks.
\newblock \emph{ArXiv}, abs/1508.05133, 2015.

\bibitem[He et~al.(2016)He, Zhang, Ren, and Sun]{res1}
He, K., Zhang, X., Ren, S., and Sun, J.
\newblock Deep residual learning for image recognition.
\newblock \emph{2016 IEEE Conference on Computer Vision and Pattern Recognition
  (CVPR)}, pp.\  770--778, 2016.

\bibitem[Henaff et~al.(2015)Henaff, Bruna, and LeCun]{graph4}
Henaff, M., Bruna, J., and LeCun, Y.
\newblock Deep convolutional networks on graph-structured data.
\newblock \emph{ArXiv}, abs/1506.05163, 2015.

\bibitem[Hinton \& Neal(1995)Hinton and Neal]{gp6}
Hinton, G.~E. and Neal, R.
\newblock Bayesian learning for neural networks.
\newblock 1995.

\bibitem[Hochreiter \& Schmidhuber(1997)Hochreiter and Schmidhuber]{lstm}
Hochreiter, S. and Schmidhuber, J.
\newblock Long short-term memory.
\newblock \emph{Neural Computation}, 9:\penalty0 1735--1780, 1997.

\bibitem[Hron et~al.(2020)Hron, Bahri, Sohl-Dickstein, and
  Novak]{attention_ntk}
Hron, J., Bahri, Y., Sohl-Dickstein, J., and Novak, R.
\newblock Infinite attention: Nngp and ntk for deep attention networks.
\newblock In \emph{ICML}, 2020.

\bibitem[Huang et~al.(2017)Huang, Liu, and Weinberger]{res2}
Huang, G., Liu, Z., and Weinberger, K.~Q.
\newblock Densely connected convolutional networks.
\newblock \emph{2017 IEEE Conference on Computer Vision and Pattern Recognition
  (CVPR)}, pp.\  2261--2269, 2017.

\bibitem[Ioffe \& Szegedy(2015)Ioffe and Szegedy]{bn}
Ioffe, S. and Szegedy, C.
\newblock Batch normalization: Accelerating deep network training by reducing
  internal covariate shift.
\newblock \emph{ArXiv}, abs/1502.03167, 2015.

\bibitem[Jacot et~al.(2018)Jacot, Gabriel, and Hongler]{NTK}
Jacot, A., Gabriel, F., and Hongler, C.
\newblock Neural tangent kernel: Convergence and generalization in neural
  networks.
\newblock In \emph{NeurIPS}, 2018.

\bibitem[Kipf \& Welling(2017)Kipf and Welling]{graph5}
Kipf, T. and Welling, M.
\newblock Semi-supervised classification with graph convolutional networks.
\newblock \emph{ArXiv}, abs/1609.02907, 2017.

\bibitem[Lecun et~al.(1998)Lecun, Bottou, Bengio, and Haffner]{conv3}
Lecun, Y., Bottou, L., Bengio, Y., and Haffner, P.
\newblock Gradient-based learning applied to document recognition.
\newblock \emph{Proceedings of the IEEE}, 86:\penalty0 2278--2324, 12 1998.
\newblock \doi{10.1109/5.726791}.

\bibitem[Lecun et~al.(2000)Lecun, Haffner, and Bengio]{conv4}
Lecun, Y., Haffner, P., and Bengio, Y.
\newblock Object recognition with gradient-based learning.
\newblock 08 2000.

\bibitem[Lee et~al.(2018)Lee, Bahri, Novak, Schoenholz, Pennington, and
  Sohl-Dickstein]{gp4}
Lee, J., Bahri, Y., Novak, R., Schoenholz, S., Pennington, J., and
  Sohl-Dickstein, J.
\newblock Deep neural networks as gaussian processes.
\newblock \emph{ArXiv}, abs/1711.00165, 2018.

\bibitem[Lee et~al.(2019)Lee, Xiao, Schoenholz, Bahri, Sohl-Dickstein, and
  Pennington]{wide}
Lee, J., Xiao, L., Schoenholz, S.~S., Bahri, Y., Sohl-Dickstein, J., and
  Pennington, J.
\newblock Wide neural networks of any depth evolve as linear models under
  gradient descent.
\newblock \emph{ArXiv}, abs/1902.06720, 2019.

\bibitem[Li \& Nguyen(2019)Li and Nguyen]{tied-encoder}
Li, P. and Nguyen, P.-M.
\newblock On random deep weight-tied autoencoders: Exact asymptotic analysis,
  phase transitions, and implications to training.
\newblock In \emph{ICLR}, 2019.

\bibitem[Littwin et~al.(2020{\natexlab{a}})Littwin, Galanti, and Wolf]{resntk}
Littwin, E., Galanti, T., and Wolf, L.
\newblock On random kernels of residual architectures.
\newblock \emph{arXiv: Learning}, 2020{\natexlab{a}}.

\bibitem[Littwin et~al.(2020{\natexlab{b}})Littwin, Myara, Sabah, Susskind,
  Zhai, and Golan]{ce}
Littwin, E., Myara, B., Sabah, S., Susskind, J., Zhai, S., and Golan, O.
\newblock Collegial ensembles.
\newblock \emph{ArXiv}, abs/2006.07678, 2020{\natexlab{b}}.

\bibitem[Matthews et~al.(2018)Matthews, Rowland, Hron, Turner, and
  Ghahramani]{gp5}
Matthews, A., Rowland, M., Hron, J., Turner, R., and Ghahramani, Z.
\newblock Gaussian process behaviour in wide deep neural networks.
\newblock \emph{ArXiv}, abs/1804.11271, 2018.

\bibitem[Novak et~al.(2019)Novak, Xiao, Bahri, Lee, Yang, Hron, Abolafia,
  Pennington, and Sohl-Dickstein]{gp7}
Novak, R., Xiao, L., Bahri, Y., Lee, J., Yang, G., Hron, J., Abolafia, D.,
  Pennington, J., and Sohl-Dickstein, J.
\newblock Bayesian deep convolutional networks with many channels are gaussian
  processes.
\newblock In \emph{ICLR}, 2019.

\bibitem[Roux \& Bengio(2007)Roux and Bengio]{gp3}
Roux, N.~L. and Bengio, Y.
\newblock Continuous neural networks.
\newblock In Meila, M. and Shen, X. (eds.), \emph{Proceedings of the Eleventh
  International Conference on Artificial Intelligence and Statistics}, volume~2
  of \emph{Proceedings of Machine Learning Research}, pp.\  404--411, San Juan,
  Puerto Rico, 21--24 Mar 2007. PMLR.
\newblock URL \url{http://proceedings.mlr.press/v2/leroux07a.html}.

\bibitem[Rumelhart et~al.(1986)Rumelhart, Hinton, and Williams]{conv5}
Rumelhart, D., Hinton, G.~E., and Williams, R.~J.
\newblock Learning internal representations by error propagation.
\newblock 1986.

\bibitem[Vaswani et~al.(2017)Vaswani, Shazeer, Parmar, Uszkoreit, Jones, Gomez,
  Kaiser, and Polosukhin]{attention2}
Vaswani, A., Shazeer, N., Parmar, N., Uszkoreit, J., Jones, L., Gomez, A.~N.,
  Kaiser, L., and Polosukhin, I.
\newblock Attention is all you need.
\newblock \emph{ArXiv}, abs/1706.03762, 2017.

\bibitem[Yang(2019{\natexlab{a}})]{yang}
Yang, G.
\newblock Tensor programs i: Wide feedforward or recurrent neural networks of
  any architecture are gaussian processes.
\newblock \emph{ArXiv}, abs/1910.12478, 2019{\natexlab{a}}.

\bibitem[Yang(2019{\natexlab{b}})]{yangScalingLimit}
Yang, G.
\newblock Scaling {{Limits}} of {{Wide Neural Networks}} with {{Weight
  Sharing}}: {{Gaussian Process Behavior}}, {{Gradient Independence}}, and
  {{Neural Tangent Kernel Derivation}}.
\newblock \emph{arXiv:1902.04760 [cond-mat, physics:math-ph, stat]}, February
  2019{\natexlab{b}}.

\bibitem[Yang(2020{\natexlab{a}})]{yang2}
Yang, G.
\newblock Tensor programs ii: Neural tangent kernel for any architecture.
\newblock \emph{ArXiv}, abs/2006.14548, 2020{\natexlab{a}}.

\bibitem[Yang(2020{\natexlab{b}})]{yang3}
Yang, G.
\newblock Tensor programs iii: Neural matrix laws.
\newblock \emph{ArXiv}, abs/2009.10685, 2020{\natexlab{b}}.

\bibitem[Yang \& Hu(2020)Yang and Hu]{yang4}
Yang, G. and Hu, E.~J.
\newblock Feature learning in infinite-width neural networks.
\newblock \emph{ArXiv}, abs/2011.14522, 2020.

\bibitem[Yang \& Schoenholz(2017)Yang and Schoenholz]{yangMFResnet}
Yang, G. and Schoenholz, S.~S.
\newblock Mean {Field} {Residual} {Network}: {On} the {Edge} of {Chaos}.
\newblock In \emph{Advances in neural information processing systems}, 2017.

\bibitem[Zou et~al.(2018)Zou, Cao, Zhou, and Gu]{zou_stochastic_2018}
Zou, D., Cao, Y., Zhou, D., and Gu, Q.
\newblock Stochastic {Gradient} {Descent} {Optimizes} {Over}-parameterized
  {Deep} {ReLU} {Networks}.
\newblock \emph{arXiv:1811.08888 [cs, math, stat]}, November 2018.
\newblock URL \url{http://arxiv.org/abs/1811.08888}.

\end{thebibliography}
\bibliographystyle{icml2020}

\appendix
\onecolumn

\paragraph{Appendix organization} The appendix is organized as follows: \\
In \cref{sec:appebdix_extra} we expands upon the examples given in \cref{sec:examples}, while adding some additional details.\\
In \cref{sec:appebdix_tp} we introduce the formal version of the \netsort{},\netsortp{} programs.\\
In \cref{sec:morediagrams} we introduce the graphical notation of \netsortp{} and demonstrate other examples of architectures or computations expressible in Tensor Programs.\\
In \cref{sec:appendix_main_proof} we prove our main result.

\paragraph{Notations}
For the readers convenience we restate the notations described in \cref{notations}, along with some additional ones which will be used throughout the appendix.
We will consider SGD with batch size 1 and learning rate of 1 (WLOG).%
We use $\xi_t$ to denote the input and $\mathcal L_t$ to denote the loss function (absorbing the label) at step $t$.
More generally, subscript $t$ on any symbol means \emph{time $t$}.
However, for brevity, we abuse notation and shorthand $f_t$ for $f_t(\xi_t)$, and, for any (pre-)activation $x$, $x_t$ for $x_t(\xi_t)$. 
We will also write $\chi_t$ for the loss derivative $\mathcal L_t'(f_t)$.
For any vector $x(\xi)$ we define $\delta x_{t+1}(\xi) \defeq \sqrt{n}\big(x_{t+1}(\xi) - x_t(\xi)\big)$ and $dx(\xi) \defeq \sqrt{n}\frac{\partial f(\xi)}{\partial x(\xi)}$.
We will track the evolution of $f$ on an arbitrary input $\tilde \xi$.%
\footnote{It might help to think of $\tilde \xi$ as some \emph{test sample}, but it can also fall in the training set.}
Similar to above, we shorthand $\tilde{x}_t,\tilde{f}_t$ for $x_t(\tilde \xi), f_t(\tilde x)$.
In general, omitting the time index $t$ for any time dependent quantity implies its value at initialization. (i.e $x(\xi) = x_0(\xi), f(\xi) = f_0(\xi)$). Finally, we use $\triangleq$ to imply equality of symbols (i.e $W^1 \triangleq W^2$ iff $W^1,W^2$ represent the same variable, as opposed to equality in value).  

\section{Additional Examples}\label{sec:appebdix_extra}
In this section we flesh out the examples given in Section 3 of the main text with the purpose of adding additional clarity, while maintaining the intuitive arguments as presented in each example to perform these calculations. The rigorous justification for these calculations will be given in the following section with the formal introduction of the Tensor Program framework.

Recall that our objective is to derive \cref{claim:NTK} by tracking the coordinate distribution of each (pre-)activations vector $x(\xi),dx(\xi) \defeq \sqrt{n}\frac{\partial f(\xi)}{\partial x(\xi)},\delta x(\xi) \defeq \sqrt{n}\big(x_{t+1}(\xi) - x_t(\xi)\big)$.
\claimNTK*

In our calculations we will rely on the following rules relating to the coordinates of any $\mathbb{R}^n$ (pre-)activation vector $x(\xi)$ in the large width regime, which we will later formalize:
\begin{itemize}
    \item $x_{t+1}(\xi) - x_t(\xi)$ has $\Theta(\frac{1}{\sqrt{n}})$ coordinates.
    \item $\delta x_{t+1}(\xi)$ has $\Theta(1)$ coordinates.
    \item $x_t(\xi) = x(\xi) + o(1)$. Consequently $Z^{x_t(\xi)} = Z^{x(\xi)}$.
    \item If $x(\xi) = \phi\big(y(\xi)\big)$ for some vector $y(\xi) \in \mathbb{R}^n$, then by Taylor approximation $\delta x_{t+1}(\xi) = \sqrt{n}\big(\phi(y_t(\xi) + \frac{\delta y_{t+1}(\xi)}{\sqrt{n}}) - \phi(y_t(\xi))\big) \approx \phi'\big(y_t(\xi)\big)\odot \delta y_{t+1}(\xi)$. Consequently $Z^{\delta x_{t+1}(\xi)} = \phi'(Z^{y_t(\xi)})Z^{\delta y_{t+1}(\xi)}$.
\end{itemize}
\begin{rem}
\label{rem:condition}
We write $Z^x$ to denote the limit coordinate distribution of $x \in \mathbb{R}^n$ conditioned on the output function $f$ at initialization. Consequently we write $\EV X^x Z^y$ to express a conditional expectation given the output function $f$. See \cref{sec:appebdix_tp} for the formal statement.
\end{rem}

\subsection{1 hidden layer}
\label{subsec:1hidden}
Recall our model is of the form:
\begin{align*}
f = V^\top x ,~~~x = \phi(h) ,~~~ h = U\xi
\end{align*}
where $\xi \in \mathbb{R}^d,~ U = \frac{u}{\sqrt{d}} \in \mathbb{R}^{n \times d},~V = \frac{v}{\sqrt{n}} \in \mathbb{R}^{n \times 1}$, for trainable parameter tensor $u$, initialized iid from $\Gaus(0,1)$, and $d=1$. 

\paragraph{Deriving The NTK}
The infinite width NTK of this architecture is given by:
\begin{align}
    \mathcal{K}(\xi,\tilde{\xi}) &=\langle\nabla_u f(\xi),\nabla_u f(\tilde{\xi})\rangle 
    = \xi^\top \tilde{\xi} \frac{dh(\xi)^\top dh(\tilde{\xi})}{n}\\
dh(\xi) &\defeq \sqrt{n}\frac{\partial f(\xi)}{\partial h(\xi)} = \phi'(h(\xi))\odot v.
\end{align}
Therefore, by LLN it follows:%
\footnote{While in \cref{rem:condition}, we said $\EV$ denotes expectation conditioned on $\lim f_0$, the NTK here does not actually depend on $\lim f_0$.}
\begin{align}\label{eqn:ntk1}
    \mathring{\mathcal{K}}(\xi,\tilde{\xi}) =  \xi^\top \tilde{\xi} \lim_{n \to \infty}\frac{dh(\xi)^\top dh(\tilde{\xi})}{n} = \xi^\top \tilde{\xi}\EV\phi'(Z^{h(\tilde{\xi})})\phi'(Z^{h(\xi)}).
\end{align}

\paragraph{Getting \cref{claim:NTK}}
To show that \cref{claim:NTK} holds with the kernel in \cref{eqn:ntk1}, we track the coordinate distribution $Z^{\delta\tilde{x}_{t+1}}$ at each step of SGD. At step $t$, the update to the weights $u_{t+1} - u_t$ is given by the gradient of the loss with respect to $u_t$:
\begin{align}
    u_{t+1} - u_t = -\chi_t\frac{ dh_t \xi_t^\top}{\sqrt{n}},~~~dh_t = \phi'(h_t)\odot v
\end{align}
Recall that $\delta\tilde{h}_{t+1} = \sqrt{n}(\tilde{h}_{t+1} - \tilde{h}_t) = \sqrt{n}(u_{t+1}\tilde{\xi} - u_t\tilde{\xi})$ and $\delta\tilde{x}_{t+1} = \sqrt{n}(\tilde{x}_{t+1} - \tilde{x}_t)$. It therefore follows:
\begin{align}
    \delta\tilde{h}_{t+1} = -\chi_t \xi_t^\top \tilde{\xi}\phi'(h_t)\odot v,~~~\delta\tilde{x}_{t+1} = \sqrt{n}\big(\phi(\tilde{h}_t + \frac{\delta\tilde{h}_{t+1}}{\sqrt{n}}) - \phi(\tilde{h}_t)\big).
\end{align}
Since $h_t = \Theta(1)$ and $\delta\tilde{h}_{t+1}=\Theta(1)$, for large $n$ we may Taylor expand $\phi$ to first order around $\tilde{h}_t$:
\begin{align}
    \delta\tilde{x}_{t+1} &\approx  \sqrt{n}\big(\phi(\tilde{h}_t) + \frac{1}{\sqrt{n}}\phi'(\tilde{h}_t)\odot  \delta\tilde{h}_{t+1} - \phi(\tilde{h}_t)\big)\\
    &=    \phi'(\tilde{h}_t)\odot \delta\tilde{h}_{t+1}\\
    &= -\chi_t\xi_t^\top \tilde{\xi}\phi'(\tilde{h}_t)\odot \phi'(h_t)\odot v \label{deltax1}.
\end{align}
Again since $\delta h_t(\xi) = \Theta(1)$, it follows that $h_t(\xi) = h(\xi) + \sum_{s=1}^{t}\frac{\delta h_s(\xi)}{\sqrt{n}} = h(\xi) + o(1)$. Hence, in the infinite width limit the coordinate distribution of $h_t(\xi)$ is identical to the coordinate distribution of $h(\xi)$ (i.e $Z^{h_t(\xi)} = Z^{h(\xi)}$). Using \cref{deltax1}, the coordinate distribution of $\delta \tilde{x}_{t+1}$ is given by:
\begin{align}
    Z^{\delta \tilde{x}_{t+1}} = -\mathring \chi_t\xi_t^\top \tilde{\xi}\phi'(Z^{\tilde{h}_t})\phi'(Z^{h_t})Z^v \label{eqn:zdeltax1}.
\end{align}
In the large width limit, the change in the output is simply given by $\tilde{f}_{t+1} - \tilde{f}_t = \frac{v^\top \delta \tilde{x}_{t+1}}{n} = \EV Z^{\delta \tilde{x}_{t+1}}Z^v$. Using \cref{eqn:zdeltax1} and the independence of $Z^v$ from the other random variables,%
\footnote{
Again, as in \cref{foot:condition}, the expectations in \cref{subsec:1hidden,subsec:2hidden} should more rigorously be interpreted as expectation conditional on the function values of  $\lim f_0$.}
\begin{align}
    \EV Z^{\delta \tilde{x}_{t+1}}Z^v &=  -\EV\mathring \chi_t\xi_t^\top \tilde{\xi}\phi'(Z^{\tilde{h}_t})\phi'(Z^{h_t})(Z^v)^2 \label{final1}\\
    &= -\mathring \chi_t\xi_t^\top \tilde{\xi}\EV\phi'(Z^{\tilde{h}_t})\phi'(Z^{h_t})\\
    &= -\mathring \chi_t\mathring{\mathcal{K}}(\xi_t,\tilde{\xi}).
\end{align}

\subsection{2 hidden layers}\label{subsec:2hidden}

Recall our model is of the form:
\begin{gather*}
f = V^\top x ,~~~x = \phi(h) ,~~~ h = W z\\
z = \phi(g) ,~~~ g = U\xi
\end{gather*}
where $U = \frac{u}{\sqrt{d}} \in \mathbb{R}^{n \times d},~W = \frac{w}{\sqrt{n}} \in \mathbb{R}^{n \times n},~V = \frac{v}{\sqrt{n}} \in \mathbb{R}^{n \times 1}$, for trainable parameters $u,w$, initialized iid from a normal distribution. As before we assume the last layer is not trained, and $d=1$. 

\paragraph{Deriving The NTK} The infinite width NTK is given by:
\begin{align}\label{eqn:ntk2}
\mathcal{K}(\xi,\tilde{\xi}) &=\langle\nabla_u f(\xi),\nabla_u f(\tilde{\xi})\rangle + \langle\nabla_w f(\xi),\nabla_w f(\tilde{\xi})\rangle\\
&= \xi^\top\tilde{\xi}\frac{dg(\xi)^\top dg(\tilde{\xi})}{n} + \frac{z(\xi)^\top z(\tilde{\xi})}{n}\frac{dh(\xi)^\top dh(\tilde{\xi})}{n}\\
    dh(\xi) &\defeq \sqrt{n}\frac{\partial f(\xi)}{\partial h(\xi)} = \phi'\big(h(\xi)\big)\odot v \label{eqn:dh2}\\
    dg(\xi) &\defeq \sqrt{n}\frac{\partial f(\xi)}{\partial g(\xi)} = \phi'\big(g(\xi)\big)\odot \big(W^\top dh(\xi)\big).\label{eqn:dg2}
\end{align}
Naively using LLN on \cref{eqn:ntk2} (and $Z^v$ being independent from everything else) should result in:
\begin{align}\label{eqn:naiveLLN}
    \mathcal{K}(\xi,\tilde{\xi}) = \xi^\top\tilde{\xi}\EV \big[Z^{dg(\xi)}Z^{dg(\tilde{\xi})}\big] + \EV \big[Z^{z(\xi)}Z^{z(\tilde{\xi})}\big]\EV\big[\phi'(Z^{h(\xi)})\phi'(Z^{h(\tilde{\xi})})\big].
\end{align}
Evaluating the term $\EV \big[Z^{dg(\xi)}Z^{dg(\tilde{\xi})}\big]$ however presents a challenge since $dh(\xi)$ depends on both $W$ and $W^\top$. As it turns out, at initialization we may naively assume that $W^\top,W$ are independent (formally known in the literature as gradient independence assumption, or GIA) \footnote{For a rigorous justification of the GIA assumption see \cite{yang2}}, we arrive using simple LLN arguments to:
\begin{align}\label{eqn:GIA}
    \EV \big[Z^{dg(\xi)}Z^{dg(\tilde{\xi})}\big] = \EV[\phi'(Z^{g(\xi)})\phi'(Z^{g(\tilde{\xi})})]\EV[\phi'(Z^{h(\xi)})\phi'(Z^{h(\tilde{\xi})})].
\end{align}
Plugging \cref{eqn:GIA} into \cref{eqn:naiveLLN} we arrive at the correct expression for the infinite width NTK. 

\paragraph{Getting \cref{claim:NTK}}
To show that \cref{claim:NTK} holds at any step $t$ (where we may not assume that GIA holds), we track the distributions of the vectors $g(\xi),z(\xi),h(\xi),x(\xi)$ throughout training.

At any step $t$ the weights are updated according to:
\begin{align}\label{eqn:weightupdate}
    u_{t+1} - u_t = -\chi_t\frac{dg_t\xi_t^\top}{\sqrt{n}},~~~ w_{t+1} - w_t = -\chi_t\frac{dh_t z_t^\top}{n}.
\end{align}

The update $\delta \tilde{g}_{t+1} \defeq \sqrt{n}(\tilde{g}_{t+1} - \tilde{g}_t),\delta \tilde{z}_{t+1} \defeq \sqrt{n}(\tilde{z}_{t+1} - \tilde{z}_t)$ are given by:
\begin{align}\label{eqn:deltag2}
    \delta \tilde{g}_{t+1} = -\chi_tdg_t\xi_t^\top \tilde{\xi},~~~\delta \tilde{z}_{t+1} = \sqrt{n}\big(\phi(\tilde{g}_t + \frac{\delta\tilde{g}_{t+1}}{\sqrt{n}}) - \phi(\tilde{g}_t))\big).
\end{align}
As before, with large $n$ we have that $\bullet_{t+1}(\xi) - \bullet_t(\xi) \sim \Theta(\frac{1}{\sqrt{n}})$ and $\delta \bullet_{t+1}(\xi) \sim \Theta(1)$ coordinates for $\bullet$ replaced by $\{g,z,h,x\}$. And so after Taylor expanding $\phi(\tilde{g}_t + \frac{\delta\tilde{g}_{t+1}}{\sqrt{n}})$ around $\tilde{g}_t$:
\begin{align}
    \delta \tilde{z}_{t+1} \approx \phi'(\tilde{g}_t)\odot \delta\tilde{g}_{t+1}. 
\end{align}

In a similar fashion, using \cref{eqn:dh2,eqn:deltag2} the updates $\delta \tilde{z}_{t+1},\delta \tilde{x}_{t+1},\delta \tilde{x}_{t+1}$ take the form:
\begin{align}
        \delta\tilde{z}_{t+1}
            &\approx \phi'(g_t)\odot \delta\tilde{g}_{t+1} =  -\chi_t\xi_t^\top \tilde{\xi}  \phi'(g_t) \odot \phi'(\tilde{g}_t) \odot \big(W^\top dh_t\big)\label{eqn:deltaz2}\\
        \delta\tilde{h}_{t+1}
            &\approx W\delta \tilde{z}_{t+1} - \chi_t \frac{z_t^\top \tilde{z}_t}{n}\phi'(h_t)\odot v  - \frac{1}{\sqrt{n}}\sum_{s=0}^t \chi_s \frac{z_s^\top \delta\tilde{z}_{t+1}}{n}\phi'(h_s)\odot v \label{eqn:deltah2}\\
            \delta \tilde{x}_{t+1} &\approx \phi'(\tilde{h}_t)\odot \delta \tilde{h}_{t+1}. \label{eqn:deltax2}
    \end{align}

where we used \cref{eqn:weightupdate} and
\begin{align}
    \delta \tilde{h}_{t+1}
    &= W_{t+1} \delta \tilde z_{t+1} + \sqrt n (W_{t+1} - W_t) \tilde z_t\\
    &= W\delta \tilde{z}_{t+1} + \sum_{s=0}^t(W_{s+1} - W_s)\delta\tilde{z}_{t+1} + \sqrt{n}(W_{t+1} - W_t)\tilde{z}_t
\end{align}
to get \cref{eqn:deltah2}. Based on \cref{eqn:dh2,eqn:deltaz2,eqn:deltah2,eqn:deltax2}, the corresponding coordinate distributions take the form:
\begin{align}
    Z^{dh_t} &= \phi'(Z^{h_t})Z^v \label{eqn:zdh2}\\
    Z^{\delta\tilde{z}_{t+1}} &= -\mathring \chi_t\xi_t^\top \tilde{\xi} \phi'(Z^{g_t})\phi'(Z^{\tilde{g}_t})Z^{W^\top dh_t}\label{eqn:zdeltag2}\\
    Z^{\delta\tilde{h}_{t+1}} &= Z^{W\delta \tilde{z}_{t+1}} - \mathring \chi_t \EV[Z^{z_t}Z^{\tilde{z}}]\phi'(Z^{h_t})Z^v\label{eqn:zdeltah2}\\
    Z^{\delta \tilde{x}_{t+1}} &= \phi'(Z^{\tilde{h}_t}) Z^{\delta \tilde{h}_{t+1}}\label{eqn:zdeltax2}.
\end{align}

As before, the functional update is given by $\lim_{n \to \infty}\tilde{f}_{t+1} - \tilde{f}_t = \lim_{n \to \infty}\frac{v^\top \delta \tilde{x}_{t+1}}{n} = \EV Z^v Z   ^{\delta \tilde{x}_{t+1}}$. Plugging \cref{eqn:zdeltax2,eqn:zdeltah2}:
\begin{align}\label{eqn:zdynamics}
    \EV Z^v Z   ^{\delta \tilde{x}_{t+1}} = -\mathring \chi_t\EV\big[Z^{z_t}Z^{\tilde{z}}\big]\EV\big[\phi'(Z^{h_t})\phi'(Z^{\tilde{h}_t})\big] - \EV\big[\phi'(Z^{h_t})Z^{W\delta \tilde{z}_{t+1}}Z^v\big].
\end{align}

To compute the second term of the RHS of \cref {eqn:zdynamics}, we use \cref{claim:Zdot}, reproduced below.
\claimZdot*

Applying \cref{claim:Zdot} to get the expression for $Z^{W\delta \tilde{z}_{t+1}}$:
\begin{align}\label{eqn:zdotform}
    Z^{W\delta \tilde{z}_{t+1}} &= G + Z^{dh_t}\EV \frac{\partial Z^{\delta\tilde{z}_{t+1}}}{\partial Z^{W^\top dh_t}}\\
    &= G -\phi'(Z^{h_t})Z^v \mathring \chi_t\xi_t^\top\tilde{\xi}\EV\big[\phi'(Z^{g_t})\phi'(Z^{\tilde{g_t}})\big]
\end{align}
As before, for $h \in \{g,z,h,x\}$ it holds that $h_t(\xi) = h(\xi) + \sum_{s=1}^t\frac{\delta h_s(\xi)}{\sqrt{n}} = h(\xi) + o(1)$, and $Z^{h_t(\xi)} = Z^{h(\xi)}$.
Plugging \cref{eqn:zdotform} into \cref{eqn:zdynamics} yields \cref{claim:NTK}.

\section{Tensor Programs: the Formal Version}\label{sec:appebdix_tp}

We briefly review the formal definition of Tensor Programs below, but readers needing more explanation and intuition should see \citep{yang3}.
We will directly describe \netsortp{} programs, which generalizes \netsort{}.

\begin{defn}\label{defn:TensorProgram}
    A \netsortp{} program is a sequence of $\R^{n}$-vectors and $\R$-scalars inductively generated via one of the following ways from an initial set $\mathcal C$ of random scalars, $\mathcal{V}$ of random $\R^{n}$ vectors, and a set $\mathcal{W}$ of random $\R^{n\times n}$ matrices (which will be sampled with iid Gaussian entries in \cref{setup:netsortplus})%
    \begin{description}
    \item [\textsc{MatMul}] same as \refMatmul{} in \cref{defn:simpleNetsorT}.
    \item [\textsc{Nonlin$^+$\label{instr:nonlin+}}] Given $\phi:\R^{k}\times\R^{l}\to\R$, previous scalars $\theta_{1},\ldots,\theta_{l}\in\R$ and vectors $x^{1},\ldots,x^{k}\in\R^{n}$, we can generate a new vector 
    \[
    \psi(x^{1},\ldots,x^{k};\theta_{1},\ldots,\theta_{l})\in\R^{n}
    \]
    where $\psi(-;\theta_{1},\ldots,\theta_{l})$ applies coordinatewise to each ``$\alpha$-slice'' $(x_{\alpha}^{1},\ldots,x_{\alpha}^{k})$. 
    \item [\textsc{Moment\label{instr:moment}}] Given same setup as above, we can also generate a new scalar 
    \[
    \f 1n\sum_{\alpha=1}^{n}\psi(x_{\alpha}^{1},\ldots,x_{\alpha}^{k};\theta_{1},\ldots,\theta_{l})\in\R.
    \]
    \end{description}
\end{defn}

A \netsort{} program is just a \netsortp{} program without scalars, without the usage of \refMoment{}, and without parameters $\theta_1, \ldots, \theta_l$ in \refNonlinp{}.

We will typically randomly sample the initial matrices, vectors, and scalars of the program as follows.
\begin{setup}\label{setup:netsortplus}
    1) For each initial $W\in\mathcal{W}$, we sample iid $W_{\alpha\beta}\sim\Gaus(0,\sigma_{W}^{2}/n)$ for some variance $\sigma_{W}^{2}$ associated to $W$, independent of other $W' \in \mathcal {W}$; 2) for some multivariate Gaussian $Z^{\mathcal{V}}=\left\{ Z^{h}:h\in\mathcal{V}\right\} \in\R^{\mathcal{V}}$, we sample the initial set of vectors $\mathcal{V}$ like $\left\{ h_{\alpha}:h\in\mathcal{V}\right\} \sim Z^{\mathcal{V}}$ iid for each $\alpha\in[n]$.
    3) For each initial scalar $\theta \in \mathcal C$, we require $\theta \asto \mathring \theta$ for some deterministic $\mathring \theta \in \R$.
\end{setup}

\newcommand{\Zz}{Z}
\newcommand{\Zhat}{\hat Z}
\newcommand{\Zdot}{\dot Z}
\newcommand{\Gg}{\mathcal G}
\newcommand{\refZhat}{\textsc{\ref{Z:Zhat}}}
\newcommand{\refZdot}{\textsc{\ref{Z:Zdot}}}
\newcommand{\refMatMul}{\textsc{\ref{instr:matmul}}}

\newcommand{\refZMatMul}{\textsc{\ref{Z:MatMul}}}
\newcommand{\refZNonlin}{\textsc{\ref{Z:Nonlin}}}
\newcommand{\refZNonlinp}{\textsc{\ref{Z:Nonlinp}}}

The following constructs a random variable $Z^h$ for every vector $h$ and a deterministic scalar $\mathring \theta$ for every scalar $\theta$ in the program.
The interpretation is that $h$ will have iid coordinates distributed like $Z^h$, and $\theta$ will converge to $\mathring \theta$ as $n \to\infty$.

\begin{defn}[$Z^h$ and $\mathring \theta$]
    \label{defn:netsortplusKeyIntuit}
Given a \netsortp{} program, we recursively define $Z^{h}$ for each vector $h$ and $\mathring{\theta}$ for each scalar $\theta$ as follows.
\begin{description}
\item [\textsc{ZInit}]
If $h\in\mathcal{V}$, then $Z^{h}$ is defined as in \cref{setup:netsortplus}. We also set $\hat{Z}^{h}\defeq Z^{h}$ and $\Zdot^{h}\defeq 0$.
\item [\textsc{ZNonlin$^+$}\label{Z:Nonlinp}] Given $\psi:\R^{k}\times\R^{l}\to\R$, previous scalars $\theta_{1},\ldots,\theta_{l}\in\R$ and vectors $x^{1},\ldots,x^{k}\in\R^{n}$, we have
\[
Z^{\psi(x^{1},\ldots,x^{k};\theta_{1},\ldots,\theta_{l})}\defeq\psi(Z^{x^{1}},\ldots,Z^{x^{k}};\mathring{\theta}_{1},\ldots,\mathring{\theta}_{l}).
\]
\item [\textsc{ZMoment}] Given same setup as above and scalar $\theta = \f 1n\sum_{\alpha=1}^{n}\psi(x_{\alpha}^{1},\ldots,x_{\alpha}^{k};\theta_{1},\ldots,\theta_{l})$, then
\[
\mathring{\theta}\defeq\EV\psi(Z^{x^{1}},\ldots,Z^{x^{k}};\mathring{\theta}_{1},\ldots,\mathring{\theta}_{l}).
\]
Here $\mathring{\theta}_{1},\ldots,\mathring{\theta}_{l}$ are deterministic, so the expectation is taken over $Z^{x^{1}},\ldots,Z^{x^{k}}$.

\item [\textsc{ZMatMul}\label{Z:MatMul}] $Z^{Wx}\defeq \hat{Z}^{Wx}+\Zdot^{Wx}$
    for every matrix $W$ (with $\Gaus(0,\sigma_{W}^{2}/n)$ entries) and vector $x$, where
    \begin{description}
    \item [\textsc{ZHat}\label{Z:Zhat}] {$\hat{Z}^{Wx}$} is a Gaussian variable with zero mean. Let $\mathcal V_W$ denote the set of all vectors in the program of the form $W y$ for some $y$.
    Then $\{\hat Z^{W y}: W y \in \mathcal V_W\}$ is defined to be jointly Gaussian with zero mean and covariance
    \[
    \Cov\left(\hat{Z}^{Wx},\hat{Z}^{W y}\right)\defeq \sigma_{W}^{2}\EV Z^{x}Z^{y},\quad\text{for any \ensuremath{Wx,W y\in\mathcal{V}_W}.}
    \]
    Furthermore, $\{\hat Z^{W y}: W y \in \mathcal V_W\}$ is mutually independent from $\{\hat Z^{v }: v \in \mathcal V \cup \bigcup_{\bar W \ne W} \mathcal V_{\bar W}\}$, where $\bar{W}$ ranges over $\mathcal{W}\cup\{A^{\trsp}:A\in\mathcal{W}\}$.
    \item [\textsc{ZDot}\label{Z:Zdot}]
    We can always unwind $Z^x = \Phi(\cdots)$, for some arguments $(\cdots)=(\{\hat Z^{W^\trsp y^i}\}_{i=1}^k, \{\hat Z^{z^i}\}_{i=1}^j; \{\mathring \theta_i\}_{i=1}^l)$, $z^i \not\in \mathcal V_{W^\trsp}$ (where $\mathcal V_{W^\trsp}$ is defined in \refZhat{}), and deterministic function $\Phi: \R^{k+j+l} \to \R$.
    Define $\partial Z^x / \partial \hat Z^{W^\trsp y^i} \defeq \partial_i \Phi(\cdots)$. 
    Then we set
    \begin{equation}
        \Zdot^{Wx}\defeq \sigma_{W}^{2}\sum_{i=1}^k Z^{y^i}\EV \f{\partial Z^x} {\partial \hat Z^{W^\trsp y^i}},\label{eqn:Zdot}
    \end{equation}
    There is some nuance in this definition, so see \cref{rem:PartialDer} and \ref{rem:ExpectationPartialDer}.
    \end{description}
\end{description}
\end{defn}

The following theorem ties the symbolic nature of the $Z$s to the analytic nature of a Tensor Program.
\begin{thm}[\netsortp{} Master Theorem, c.f.\ Theorem E.15 of \citep{yang3}]
    \label{thm:PLNetsorT+MasterTheorem} Fix a Tensor Program initialized accordingly to \cref{setup:netsortplus}.
    Adopt \cref{assm:MasterTheoremSmoothness}.
    Then 
    \begin{enumerate}
    \item For any fixed $k$ and any pseudo-Lipschitz $\psi:\R^{k}\to\R$, as $n\to\infty$,
    \begin{equation}
    \f 1n\sum_{\alpha=1}^{n}\psi(h_{\alpha}^{1},\ldots,h_{\alpha}^{k})\asto\EV\psi(Z^{h^{1}},\ldots,Z^{h^{k}}),
    \label{eqn:mastertheorem}
    \end{equation}
    for any vectors $h^{1},\ldots,h^{k}$ in the program, where $Z^{h^{i}}$ are as defined in \cref{{defn:netsortplusKeyIntuit}}.
    \item Any scalar $\theta$ in the program tends to $\mathring{\theta}$ almost surely, where $\mathring{\theta}$ is as defined in \cref{{defn:netsortplusKeyIntuit}}.
    \end{enumerate}
\end{thm}

\begin{rem}[Partial derivative]\label{rem:PartialDer}
    The partial derivative in \refZdot{} should be interpreted as follows.
    By a simple inductive argument, $Z^x$ for every vector $x$ in the program is defined \emph{uniquely} as a deterministic function $\varphi(\Zhat^{x^1}, \ldots, \Zhat^{x^k})$ of some $x^1, \ldots, x^k$ in $\mathcal V$ or introduced by \refMatMul{} (notationally, we are suppressing the possible dependence on limit scalars $\mathring \theta_1, \ldots, \mathring \theta_l$).
    For instance, if in a program we have $A \in \mathcal W, v \in \mathcal V$, $y = A v, x = A^\trsp y$, then $Z^x = \Zhat^x + \Zhat^v$, so $\varphi$ is given by $\varphi(a, b) = a + b$.
    Then
    \begin{equation*}
        \partial Z^x / \partial \Zhat^{x^i} \defeq \partial_i \varphi(\Zhat^{x^1}, \ldots, \Zhat^{x^k}), \quad \text{and} \quad \partial Z^x / \partial \Zhat^z \defeq 0 \text{ for any } z \not\in\{x^1, \ldots, x^k\}.
    \end{equation*}
    Note this definition depends on the precise way the program is written, not just on the underlying mathematics.
    For example, if $y,z \in \mathcal V$ and $x = \phi(W(y+z))$, then $Z^x = \phi(\Zhat^{W(y+z)})$ so that $\partial Z^x/\partial \Zhat^{Wy} = \partial Z^x/\partial \Zhat^{Wz} = 0$.
    If instead, we have $x = \phi(Wy+Wz)$, then $Z^x = \phi(\Zhat^{Wy} + \Zhat^{Wz})$ so that $\partial Z^x/\partial \Zhat^{W(x+y)} = 0$.
    However, in both cases, $\Zdot^{W^\trsp x} = (Z^y + Z^z)\EV \phi'(\Zhat^{W(y+z)})$.
\end{rem}

\begin{rem}[Partial derivative expectation]\label{rem:ExpectationPartialDer}
    The quantity $\EV\frac{\partial Z^{x}}{\partial\hat{Z}^{W^{\trsp}y}}$ is well defined if $Z^{x}$ is differentiable in $\hat{Z}^{W^{\trsp}y}$. However, even if this is not the case, e.g. if $x=\theta(W^\trsp y)$ where $\theta$ is the Heavyside step function, we can still define this expectation by leveraging Stein's lemma:
    
    In \refZdot{}, suppose $\{W^{\trsp}y^{i}\}_{i=1}^{k}$ are all elements of $\mathcal V_{W^\trsp}$ introduced before $x$.
    Define the matrix $C\in\R^{k\times k}$ by $C_{ij}\defeq \EV Z^{y^{i}}Z^{y^{j}}$ and define the vector $b\in\R^{k}$ by $b_{i}\defeq \EV\hat{Z}^{W^{\trsp}y^{i}}Z^{x}$. If $a=C^{+}b$ (where $C^{+}$ denotes the pseudoinverse of $C$), then in \refZdot{} we may set
    \begin{equation}
    \sigma_{W}^{2}\EV\frac{\partial Z^{x}}{\partial\hat{Z}^{W^{\trsp}y^{i}}}=a_{i}.\label{eq:ExpectationPartialDer}
    \end{equation}
    This definition agrees with the partial derivative expectation by Stein's lemma when the latter is well defined.
    \cref{thm:PLNetsorT+MasterTheorem} holds with this broader definition of partial derivative expectation.
\end{rem}

\paragraph{Pseudo-Lipschitz functions}

are, roughly speaking, functions whose weak derivatives are polynomially bounded.
\begin{defn}\label{defn:pseudoLipschitz}
    A function $f: \R^k \to \R$ is called \emph{pseudo-Lipschitz} of degree $d$ if $|f(x) - f(y)| \le C\|x-y\|(1 + \sum_{i=1}^k |x_i|^d + |y_i|^d)$ for some $C$.
    We say $f$ is pseudo-Lipschitz if it is so for any degree.
\end{defn}
Here are some basic properties of pseudo-Lipschitz functions:
\begin{itemize}
    \item The norm $\|\cdot\|$ in \cref{defn:pseudoLipschitz} can be any norm equivalent to the $\ell_2$ norm, e.g. $\ell_p, p\ge1,$ norms. Similarly, $\sum_{i=1}^k |x_i|^d + |y_i|^d$ can be replaced by $\|x\|^d_p + \|y\|^d_p$, for any $p \ge 1$.
    \item A pseudo-Lipschitz function is polynomially bounded.
    \item A composition of pseudo-Lipschitz functions of degrees $d_1$ and $d_2$ is pseudo-Lipschitz of degree $d_1 + d_2$.
    \item A pseudo-Lipschitz function is Lipschitz on any compact set.
\end{itemize}

We adopt the following assumption for the Master Theorem \cref{thm:PLNetsorT+MasterTheorem}.
\begin{assm}
    \label{assm:MasterTheoremSmoothness}
    Suppose
    \begin{enumerate}
        \item If a function $\phi(;-): \R^{0 + l} \to \R$ with only parameter arguments is used in \refMoment{}, then $\phi$ is continuous in those arguments.
        \item Any other function $\phi(-;-): \R^{k + l} \to \R$ with parameters (where $k>0$) used in \refNonlin{} or \refMoment{} is pseudo-Lipschitz in all of its arguments (both inputs and parameters).
    \end{enumerate}
\end{assm}
Statement 1 in \cref{assm:MasterTheoremSmoothness} essentially says that if we have scalars $\theta_1, \ldots, \theta_l$ in the program, then we can produce a new scalar by applying a continuous function (a weaker restriction than a pseudo-Lipschitz function)  to them.
Indeed, if $\theta_1, \ldots, \theta_l$ converge almost surely, then this new scalar does too.
In our setting, statement 1 is used to allow any loss function whose derivative is continuous.

Other versions of the Master Theorem can be found in \citep{yang3}, for example, versions where the we do not assume any smoothness condition at all on the nonlinearities beyond that they be polynomially bounded, in exchange for assuming what's called a \emph{rank stability} condition.
This rank stability should be generically true, but checking it rigorously is subtle, so we are content with the pseudo-Lipschitz condition in this paper.

\section{More Diagrams}
\label{sec:morediagrams}

\begin{figure}
    \centering
    \includegraphics[width=\textwidth]{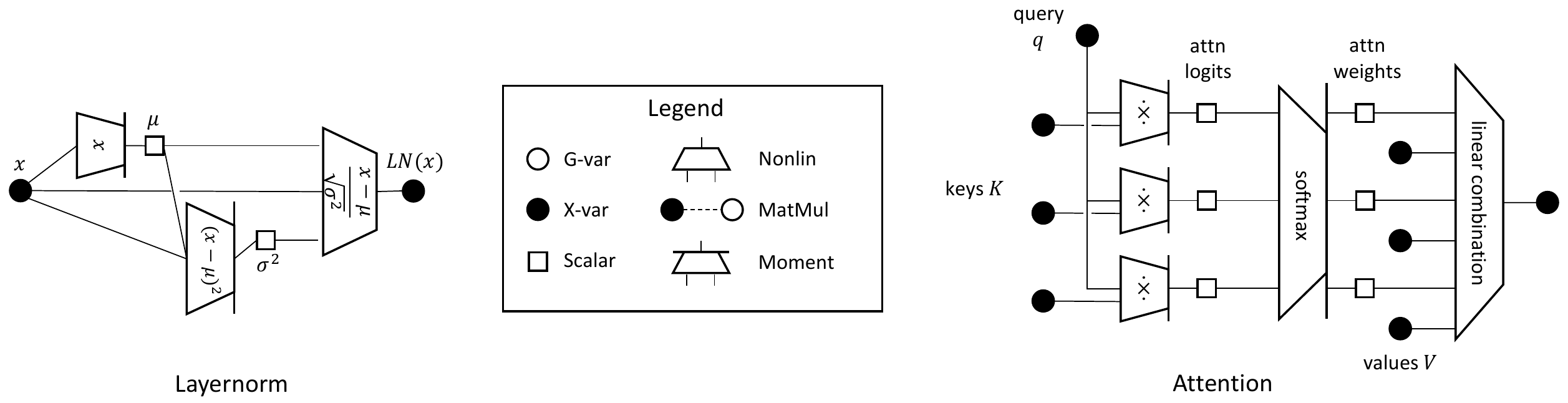}
    \caption{Layernorm and attention can be implemented with \netsortp{}.}
    \label{fig:ln_attn}
\end{figure}

We can augment the graphical form of \netsort{} to accomodate the \refMoment{} instruction in \netsortp{}.
See \cref{fig:ln_attn} for an example for layernorm and attention.
In short, we denote scalar variables with a square, in contrast to the circle for vector variables, and we use a ``bar-gate'' to denote the \refMoment{}, where the function in the gate corresponds to $\psi$ in \refMoment{}.

\begin{figure}
    \centering
    \includegraphics[width=0.5\textwidth]{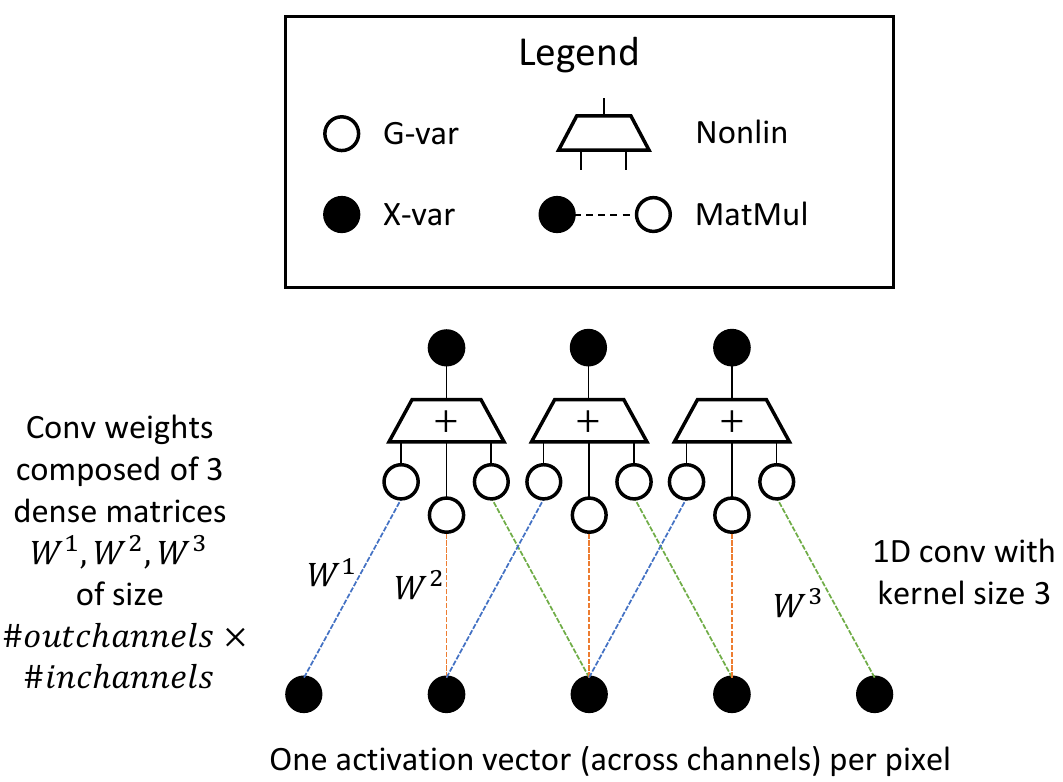}
    \caption{Convolution can be implemented with \netsort{}.}
    \label{fig:conv}
\end{figure}

\begin{figure}
    \centering
    \includegraphics[width=0.4\textwidth]{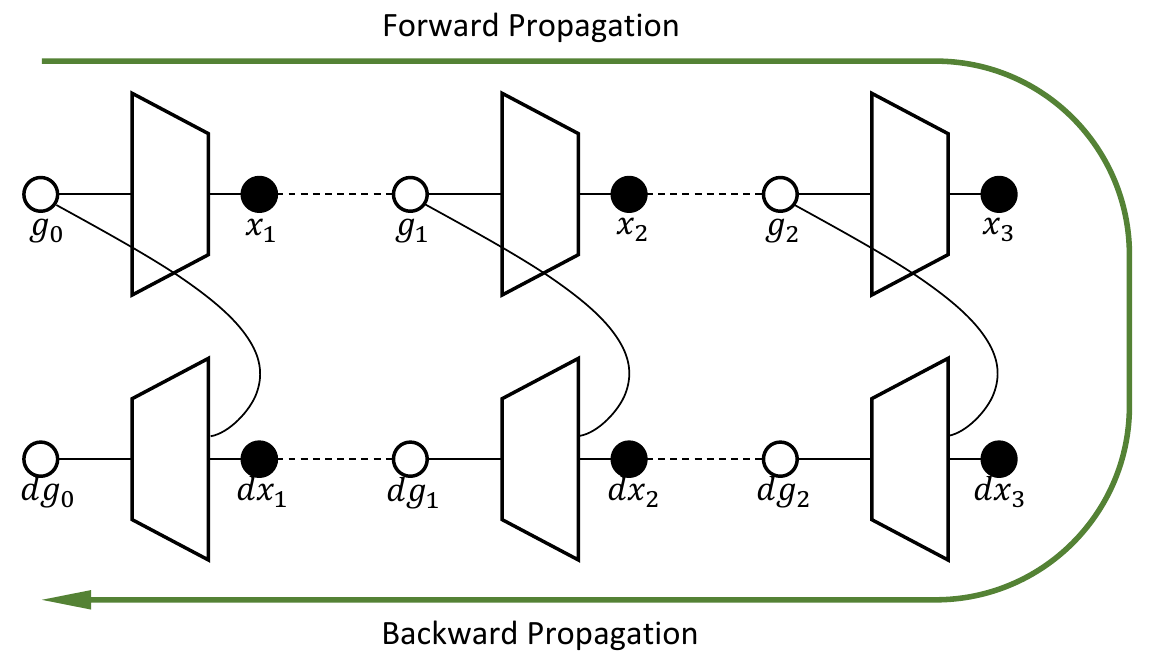}
    \caption{Backpropagation (of an MLP $f$) can be implemented in \netsort{}.
    However, \netsort{} cannot express the loss derivative, so we cannot unroll multiple steps of SGD in \netsort{}, unlike \netsortp{}.}
    \label{fig:backprop}
\end{figure}

In addition, for more examples of the expressivity of \netsort{}, \cref{fig:conv,fig:backprop} demonstrate convolution and MLP backpropagation in \netsort{}.

\section{Proof of Main Result}\label{sec:appendix_main_proof}
We dedicate the following section to prove \cref{thm:main}. We will begin by proving a simplified version under the same assumptions as \cref{sec:proofsketch}, as reproduced below:
\newlist{assumptions}{enumerate}{10}
\setlist[assumptions]{label*=A\arabic*.}

\begin{setup}[Representable NN in NTK Parametrization]\label{setup:main_simplified}
    Suppose a neural network $f \in \mathbb{R}$ is represented by a \netsort{} program (in the sense of \cref{defn:nnrep}) whose \refNonlin{} all have polynomially bounded derivatives.%
    \footnote{More generally, we can allow any pseudo-Lipschitz function here, but for simplicity we go with the statement in the main text.}
    Adopt the NTK parametrization: for every matrix parameter $W \in \R^{n\times n}$ of $f$, we factor $W = \f 1 {\sqrt n} w$ where $w$ is the trainable parameter;
    likewise, for each input layer matrix $U^i \in \R^{n\times d}$, we factor $U^i = \f 1 {\sqrt d} u^i$,
    and likewise the output matrix $V = \f 1 {\sqrt n} v$. 
    We randomly initialize all trainable parameters iid as $\Gaus(0, 1)$. 
    Furthermore, we assume the following:
    \begin{assumptions}
        \item Input and output layers $\{u^i\},v$, as well as biases are not trained (only $\mathbb{R}^{n \times n}$ weight matrices are trained)\label{ass:first_last}.
        \item The forward pass does not use both a matrix and its transpose (in different \refMatMul{}s).\label{ass:trans}
        \item The output is a scalar $f \in \mathbb{R}$.\label{ass:scalar}
        \item We assume the last layer embedding is a G-var. \label{ass:GvarEmbedding}
    \end{assumptions}
\end{setup}

Our main result is to show that the SGD training of such a neural network described in \cref{setup:main} reduces to kernel gradient descent with kernel $\mathring{\mathcal K}$ in the infinite-width limit.

\begin{thm}[\ref{NTKtrain} is Architecturally Universal]\label{thm:main_simplified}
    Consider training a network $f$ described in \cref{setup:main_simplified} via SGD with batch-size 1 and (WLOG) learning rate 1.
    Let $\xi_t$ be the input and $\loss_t: \R \to \R$ be the loss function (absorbing the label) at time $t$.
    Suppose $\loss_t$ is continuous for all $t$.
    Then, for any $\xi$ and $t$, $f_t(\xi)$ converges almost surely to a random variable $\mathring f_t(\xi)$ as width $\to \infty$, such that
    \begin{align}%
        \mathring f_{t+1}(\xi) - \mathring f_t(\xi) = - \mathring{\mathcal K}(\xi, \xi_t) \loss_t'(\mathring f_t(\xi_t))
    \end{align}
    where $\mathring{\mathcal{K}}$ is the infinite-width NTK (at initialization) of the neural network.
    
    \end{thm}

\renewcommand{\bar}{\mathtt}

\subsection{SGD as a \netsortp{} Program}
SGD is comprised of a sequence of forward and backward passes computed on some architecture. WLOG, let $\pi_0$ denote the \emph{reduced program} implementing the body of network $f$, and let $x(\xi)$ denote the final embedding such that $f(\xi) = V^\top x(\xi)$, we will now show how the SGD procedure on $\pi_0$ can be implemented by a \netsortp{} program.  
\subsubsection{First Forward Pass} \label{subsub:first_pass}

While $\pi_0$ implements the embeddings $x(\xi)$ by definition, the outputs $f(\xi)$ cannot be implemented trivially in a program since that at initialization $f(\xi) = \frac{v^\top x(\xi)}{\sqrt{n}}$ is not deterministic, and converges non-trivially to a GP, violating the requirements of a scalar type in a \netsortp{} program which require all scalar types to converge to a deterministic limit as $n \to \infty$.
Nevertheless, we can still easily express evolution of $f$ \emph{conditioned on (i.e.\ fixing) the values of $f$ at initialization}.
More formally, let $\bar{f} = [f(\xi_0),f(\xi_1),...,f(\xi_{D-1})]^\top \in \mathbb{R}^{D}$ denote a fixed vector of outputs, and let $X  = [x(\xi_0),x(\xi_1),...,x(\xi_{D-1})]^\top \in \mathbb{R}^{n \times D}$ denote a fixed embedding matrix such that $\bar{f} = \frac{X^\top v}{\sqrt{n}}$. The distribution of $v$ when conditioned on $\bar{f}$ and $X$ is given by (see e.g.\ \citep[Sec K.2]{yang3})
    \begin{equation}        
    v \disteq_{\bar f, X} \sqrt{n} X^+\bar{f} + \Pi \bar v
    \label{eqn:v}
    \end{equation}
where $X^+$ is the pseudo-inverse of $X$, $\bar v$ is an independent copy of $v$ and $\Pi$ is the projection operator projecting unto the  orthogonal complement of the space spanned by $X$. Namely:
    \[
    X^+ = \frac{1}{n}X(\frac{X^\top X}{n})^{+},\quad\Pi = I - X^+ X^\top
    \]
Denote $\Sigma = \frac{X^\top X}{n} \in \R^{D \times D}, \mu = \frac{X^\top v}{n} \in \R^{D}$.
Define
    \[
    \hat v \defeq X (\frac{\Sigma^+ \bar{f}}{\sqrt{n}})  + \bar v - X\Sigma^+ \mu.
    \label{eqn:hatv}
    \]
Then we see via \cref{eqn:v} that
\[
    v \disteq_{\bar f, X} \hat v.\]
Given $\bar v$ and (the columns of) $X$ as vectors and $\bar f$ as scalars in a program, $\hat v$ may be defined in the same program via \refNonlin{}, where $\frac{\Sigma^+ \bar{f}}{\sqrt{n}}$ and $\Sigma^+ \mu$ (both finite-dimensional) provide coefficients for the linear combination over (columns of) $X$.
Formally, to express the evolution of $f$ conditioned on $f_0 = \bar f$ at initialization, the program will calculate the first forward pass up to $X$, calculate the loss derivatives $\chi$ assuming $f_0 = \bar f$, and then proceed with the backward pass and later forward/backward passes with $v$ replaced by $\hat v$.

However, since $\frac{\Sigma^+ \bar{f}}{\sqrt{n}}, \mu \asto 0$ and $\Sigma^+ \asto \mathring \Sigma^+$ (by rank stability, c.f.\ \citep[Lemma L.11]{yang3}), these coefficients of the linear combination converge to 0, so that
$Z^{\hat v} = Z^{\bar v}$.
Intuitively, this means that the distribution of $v$ conditioned on the equality $\bar f = X^\trsp v/\sqrt n$ is asymptotically the same as no conditioning as $n \to \infty$.
Thus, for the limit calculation of $\delta f_t$ and other quantities, it ends up not mattering whether we use $\hat v$ or $v$.

\paragraph{Computing The Loss Derivatives} The loss derivative $\chi (\xi) = \frac{\partial \mathcal{L}(f(\xi))}{\partial (f(\xi))}$ after the first forward pass given $f(\xi)$ can be implemented with \refMoment{} instructions using $\psi(;f(\xi)) = \mathcal{L}'(f(\xi))$.

\subsubsection{Implementing SGD}\label{subsec:unrolled_SGD}
Under SGD, the update at step $t+1$ to any weight $w \in \mathbb{R}^{n \times n}$ is given by:
\begin{align}
    w_{t+1} - w_t &= -\chi_t\sum_{\bar{g},\bar{h}:\bar{g} = W\bar{h}}\frac{d\bar{g}_t\bar{h}_t^\top}{n}  \label{eqn:w_update}.
\end{align}
where the summation in \cref{eqn:w_update} is over all pairs of vectors $\bar{g},\bar{h}$ in program $\pi_0$ satisfying $\bar{g} = W\bar{h}$ (there can be multiple such pairs since $\pi_0$ may reuse the same matrix $W$).

To write the full unrolled SGD as a \netsortp{} program, we will need to implement the error signal $dg_t \defeq \sqrt{n}\frac{\partial f_t}{\partial g_t}$ for each G-var $g$ at time $t$. To accomplish this, we recall the notion of paths in program $\pi_0$:
\Paths*
Note that a path $p$ represents a series of nodes independent of an input, and can be instantiated as $p(\xi)$ by an input $\xi$, resulting in a series of instantiated G-vars and X-vars $p^i(\xi)$.

For any G-var $g =Wh$, we can write the error term $dg$ as the summation of errors signals over paths $p$: 
\begin{align}\label{eqn:error+paths}
dg(\xi) &=  \sum_{p:p^{-1} = x,p^{1} = g} J^p(\xi)\\
\text{where}\quad J^p &= (\frac{\partial p^{2}}{\partial p^1})^\top  (\frac{\partial p^{-2}}{\partial p^{-3}})^\top...(\frac{\partial p^{-1}}{\partial p^{-2}})^\top v
\end{align}
(Here again, $J^p$ represents a symbolic computation that can be instantiated with an input $J^p(\xi)$). Note that $J^p$ can be defined recursively:
\begin{align}\label{eqn:rec}
    J^p = (\frac{\partial p^{2}}{\partial p^1})^\top (\frac{\partial p^{3}}{\partial p^2})^\top J^{p:3} 
\end{align}
where $J_{p:k},k\leq|p|$ is defined as:
\[
J^{p:k} \defeq \begin{cases} 
   (\frac{\partial p^{k+1}}{\partial p^k})^\top  (\frac{\partial p^{-2}}{\partial p^{-3}})^\top...(\frac{\partial p^{-1}}{\partial p^{-2}})^\top v & k<|p|\\
    v & k=|p|
    \end{cases}
\]
Recall that each path $p$ starts with an X-var $p^0$, and alternates between G and X vars. Let $W^{p^3}$ denote the defining weight matrix of G-var $p^3$ (i.e $p^3 = W^{p^3}p^2$), and let $p^2 = \psi(...,p^1,...)$. Then we can re-write \cref{eqn:rec} as:
\begin{align}\label{eqn:path_der}
J^p = \psi'(...,p^1,...) \odot J^{p:2},~~~ J^{p:2} = (W^{p^3})^\top J^{p:3}
\end{align}
Note that \cref{eqn:path_der} can be written in \netsort{} language using \refMatmul{} instructions using the transposed weights, and \refNonlin{} instructions using $\psi'$, which is pseudo-Lipschitz by \cref{setup:main_simplified}.

Recall that $\pi_0$ is the program defining the network architecture.
We now write the unrolled SGD of this network in a new program $\pi$.
Below, recall that lack of time subscript means $t=0$ (e.g.\ $W$ means $W_0$, the initialized value).
In addition, feel free to revisit the notations explained before \cref{sec:appebdix_extra}.
\begin{itemize}
    \item If $g = Wh \in \pi_0$, then:
    \begin{align}
        \delta \tilde{g}_{t+1} &= \sqrt{n}(W_{t+1}\tilde{h}_{t+1} - W_t\tilde{h}_t)
        = W\delta \tilde{h}_{t+1} + \sqrt{n}(W_{t+1} - W_t)\tilde{h}_t + \sum_{s=0}^t (W_{s+1} - W_s)\delta \tilde{h}_{t+1}. \label{eqn:deltag_allterms}
    \end{align}
    where, using \cref{eqn:w_update}, we have
    \begin{align}
        \sqrt{n}(W_{t+1} - W_t)\tilde{h}_t &= - \chi_t\sum_{\bar{g},\bar{h}:\bar{g} = W\bar{h}}d\bar{g}_t\frac{\bar{h}_t^\top \tilde{h}_t}{n} \label{eqn:SGD_TERM1}\\
        \sum_{s=0}^t (W_{s+1} - W_s)\delta \tilde{h}_{t+1}&= -\sum_{s=0}^t\frac{\chi_s}{\sqrt{n}}\sum_{\bar{g},\bar{h}:\bar{g} = W\bar{h}}d\bar{g}_s\frac{\bar{h}_s^\top \delta\tilde{h}_t}{n}\label{eqn:SGD_TERM2}
    \end{align}
    
    \paragraph{Tensor Program implementation} \cref{eqn:deltag_allterms,eqn:SGD_TERM1,eqn:SGD_TERM2} may be easily implemented using \netsortp{} instructions. 
    For instance, \cref{eqn:SGD_TERM1} (assuming the sum sums over a single pair $\{\bar{h},\bar{g}\}$) may be implemented using \refMoment{} and \refNonlinp{} instructions as follows: the term $\frac{\bar{h}_t^\top \tilde{h}_t}{n}$ may be implemented by a \refMoment{} instruction with $\psi(\tilde{h}_t,\bar{h}_t) = \frac{1}{n}\sum_\alpha (\tilde{h}_t)_\alpha(\bar{h}_t)_\alpha$. The full term is then a \refNonlinp instructions $\psi(d\bar{g}_t;\chi_t,\{\frac{\bar{h}_t^\top \tilde{h}_t}{n}\}_{\bar h})$ with scalars $\chi_t, \{\frac{\bar{h}_t^\top \tilde{h}_t}{n}\}_{\bar h}$ and vector $d\bar{g}_t$.

    \item If $g = \psi(h^1,...,h^k) \in \pi_0$, then
    \begin{align}
        \delta \tilde{g}_{t+1} = \sqrt{n}\left(\psi (\tilde{h}_t^1 + \frac{\delta\tilde{h}_{t+1}^1}{\sqrt{n}},...,\tilde{h}_t^k + \frac{\delta\tilde{h}_{t+1}^k}{\sqrt{n}}) - \psi(\tilde{h}_t^1,...,\tilde{h}_t^k)\right).\label{eqn:delta_nonlin}
    \end{align}
    
    \paragraph{Tensor Program implementation} \cref{eqn:delta_nonlin} may be implemented as a \refNonlinp{} instruction:
    \[\delta \tilde{g}_{t+1} := \psi^\star \left(\{\tilde{h}_t^i\}_{i=1}^k \cup \{\delta\tilde{h}_{t+1}^i\}_{i=1}^k;\frac{1}{\sqrt{n}}\right)\] for a set of vectors $\{\tilde{h}_t^i\}_{i=1}^k,\{\delta\tilde{h}_{t+1}^i\}_{i=1}^k$ and a scalar $\frac{1}{\sqrt{n}}$, where:

    \begin{align}\label{eqn:psistar}
        \psi^\star(\{\mu^i\}_{i=1}^k \cup \{\nu^i\}_{i=1}^k;\theta)_\alpha \defeq \begin{cases}
        \frac{\psi(\mu^1 + \theta \nu^1,...,\mu^k + \theta \nu^k)_\alpha - \psi(\mu^1 ,...,\mu^k)_\alpha}{\theta} & \theta>0\\
        \sum_{i=1}^k \frac{\partial \psi(\mu^1 ,...,\mu^k)_\alpha}{\partial \mu^i_\alpha}\nu^i_\alpha & \theta=0. 
        \end{cases}
    \end{align}
    Since $\psi'$ is pseudo-Lipschitz by \cref{setup:main_simplified}, $\psi^\star$ is pseudo-Lipschitz in all of its inputs as well.
    \item WLOG assume $f(\xi) = \frac{v^\top x(\xi)}{\sqrt{n}}$, then:
    \begin{align}
        \tilde{f}_{t+1} = \frac{v^\top \delta\tilde{x}_{t+1}}{n},~~~\chi_t = \nabla_{f_t}\mathcal{L}_t.\label{eqn:scalars}
    \end{align}
    \paragraph{Tensor Program implementation} the scalar type outputs $f_t(\xi)$ at $t>0$ for any input $\xi$ can be implemented using the \refMoment{} instruction. The loss derivative $\chi_t,t>0$ given $f_t$ can be implemented with \refMoment{} instructions using $\psi(-;f(\xi)) = \mathcal{L}'\big(f(\xi)\big)$ where $f(\xi)$ is treated as a scalar type as in the first forward pass.

    \item If $g \in \mathbb{R}^n \in \pi_0$:
    \begin{align}\label{eqn:g_update}
        g_{t+1}(\xi) = g(\xi) + \frac{1}{\sqrt{n}}\sum_{s=1}^{t+1}\delta g_s(\xi)
    \end{align}
    \paragraph{Tensor Program implementation} $\tilde{g}_{t+1}$ is implemented using a \refNonlinp{} instruction $g_{t+1}(\xi) = \psi \big(g(\xi)\cup\{\delta g_s(\xi)\}_{s=1}^{t+1};\frac{1}{\sqrt{n}}\big)$ with $\psi(\mu\cup\{\nu_s\}_{s=1}^{t+1};\theta) = \mu + \theta\sum_s\nu_s$.
    
    \item If $g = Wh \in \pi_0$, then using \cref{eqn:error+paths}:
    \begin{align}\label{eqn:dupfate1}
        dg(\xi)_{t+1} &=  \sum_{p:p^{-1} = x,p^{1} = g} J^p_{t+1}(\xi)\\
        J^p_{t+1} &= \psi'(...,p_{t+1}^1,...) \odot \big((W_{t+1}^{p^3})^\top J^{p:3}_{t+1}\big)        
    \end{align}
    Using \cref{eqn:rec}, $J^p_{t+1}$ is implemented recursively starting from $J^{p:-2}_{t+1} = (W_{t+1}^{p^{-1}})^\top v$. Plugging in the weights update at time $t+1$ (\cref{subsec:unrolled_SGD}):
    \begin{align}
    J^{p:-2}_{t+1} &= (W_t^{p^{-1}} + W_{t+1}^{p^{-1}} - W_t^{p^{-1}})^\top v = J^{p:-2}_t -\chi_t\sum_{\bar{g},\bar{h}:\bar{g} = W_{t+1}^{p^{-1}}\bar{h}}\frac{\bar{h}_t}{\sqrt{n}}\frac{ d\bar{g}_t^\top v}{n} \label{eqn:dupfate2}\\
    J^p_{t+1} &= \psi'(...,p_t^1 + \frac{\delta p_{t+1}^1}{\sqrt{n}},...) \odot J^{p:2}_{t+1}\label{eqn:dupfate3}
    \end{align}
    \paragraph{Tensor Program implementation} $dg(\xi)_{t+1}$ is implemented using \refMoment{} and \refNonlinp{} instructions.

\end{itemize}

\begin{figure}
    \centering
    \includegraphics[width=0.8\textwidth]{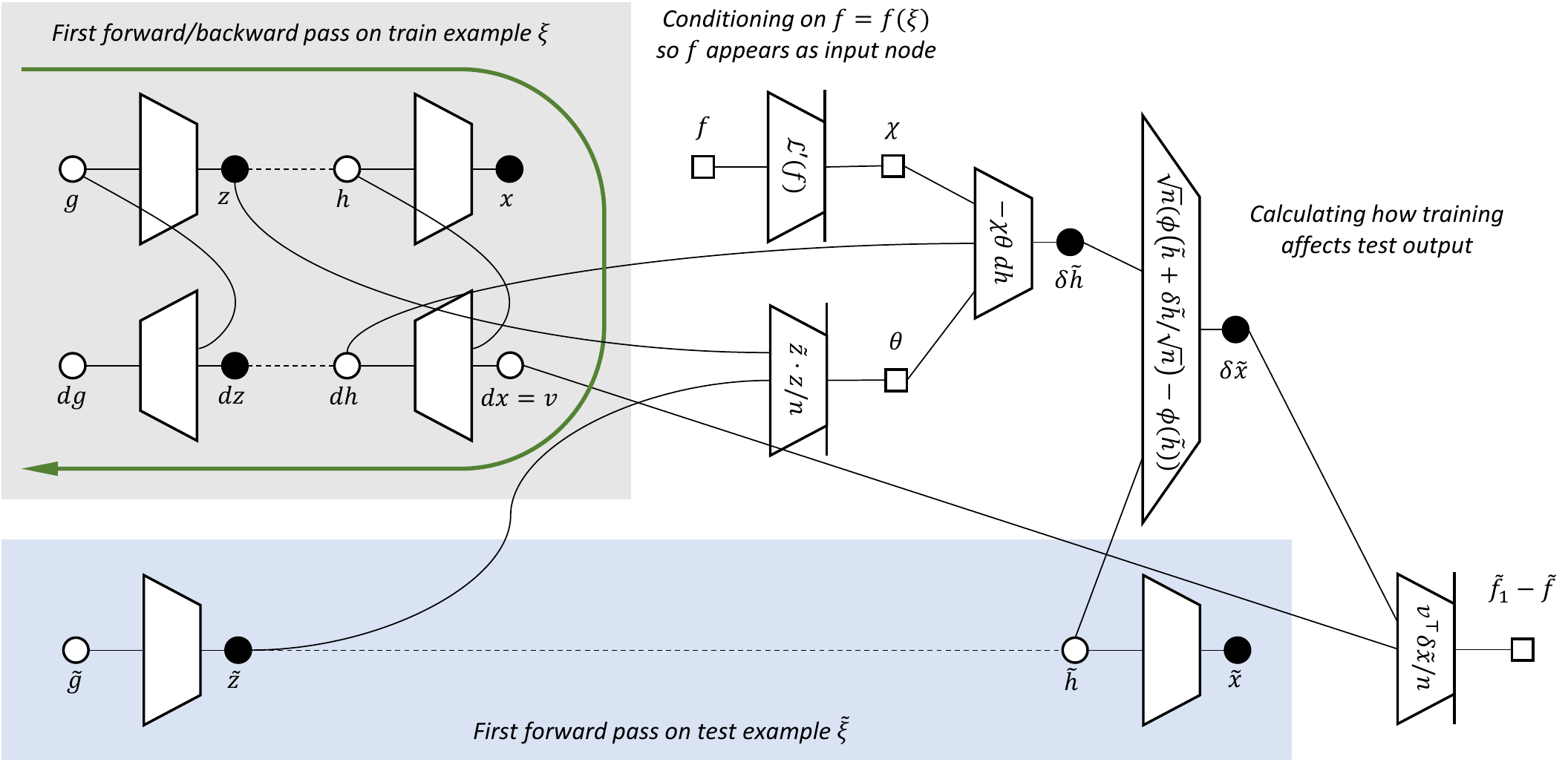}
    \caption{Graphical form of \netsortp{} program (\cref{alg:example}) encoding the first training step and the effect on the test set  on the two hidden layer MLP specified in \cref{subsec:2hidden}, but where only the middle layer $w$ is trained.
    Technically, we should have $\hat v$ as defined in \cref{eqn:hatv} instead of $v$, but as explained there, this does not affect the limit.}
    \label{fig:firsstepprogram}
\end{figure}

For further illustration, we present in \cref{alg:example} a program implementation of the first iteration of SGD on the two hidden layer MLP specified in \cref{subsec:2hidden}, but where only the middle layer $w$ is trained; in \cref{fig:firsstepprogram} we have its graphical form.
Naturally, the following SGD steps can be implemented in a similar fashion.

\begin{algorithm}[h]%
   \caption{\netsortp{} program $\pi_1$ implementing the first update $\tilde{f}_1 - \tilde{f}$. Note that in this example, $\psi$ represents multiple functions, while $\phi$ is a fixed function representing the non-linearity of the MLP in \cref{subsec:2hidden}.
   Technically, we should have $\hat v$ as defined in \cref{eqn:hatv} instead of $v$, but as explained there, this does not affect the limit.}
   \label{alg:example}
\begin{algorithmic}
   \STATE {\bfseries Input: $\mathcal{W} = \{w\},\mathcal{V} = \{v,g = g_0(\xi_0),\tilde{g} = g_0(\tilde{\xi})\},\mathcal{C} = \{f = f_0(\xi_0),\tilde{f} = f_0(\tilde{\xi})\}$}\\
   \refMoment{}: $\chi:= \mathcal{L}'(f)$\\
   \refNonlinp{}: $z:= \phi (g)$\\
   \refNonlinp{}: $\tilde{z}:= \phi (\tilde{g})$\\
   \refMatmul{}: $h:= Wg$\\
   \refMatmul{}: $\tilde{h}:= W\tilde{g}$\\
   \refNonlinp{}: $\tilde{z}:= \phi (\tilde{g})$\\
   \refMoment{}: $\theta:= \psi(z,\tilde{z}) = \frac{z^\top \tilde{z}}{n}$\\
   \refNonlin{}: $dh:=\psi(v,h) = \phi'(h)\odot v$\\
   \refNonlinp{}: $\delta\tilde{h}_1:= \psi(dh;\theta,\chi) = -\chi \theta dh$\\
   \refNonlinp{}: $\delta\tilde{x}:= \psi(\tilde{h},\delta\tilde{h}_1;\frac{1}{\sqrt{n}}) = \sqrt{n}\big(\phi(\tilde{h} + \frac{\delta\tilde{h}_1}{\sqrt{n}}) - \phi(\tilde{h})\big)$\\
   \refMoment{}: $\tilde{f}_1 - \tilde{f} = \frac{v^\top \delta\tilde{x}}{n}$\\
   \STATE{\bfseries Output: $\tilde{f}_1 - \tilde{f}$}
\end{algorithmic}
\end{algorithm}

\paragraph{SGD in the Infinite Width Limit}

According to the \netsortp{} rules as specified in \cref{defn:netsortplusKeyIntuit}, we have the following identities:
\begin{itemize}
\item If $g = Wh$, then using \cref{eqn:deltag_allterms,eqn:SGD_TERM1,eqn:SGD_TERM2}:
 (Here $Z^{d \bar g_t} = Z^{d \bar g_t(\xi_t)}$, $Z^{\bar h_t} = Z^{\bar h_t (\xi_t)}$, and $Z^{\tilde h} = Z^{h_t(\tilde \xi)}$)
\begin{align}
    Z^{\delta \tilde{g}_{t+1}} &= Z^{W\delta\tilde{h}_{t+1}} - \chi_t\sum_{\bar{g},\bar{h}:\bar{g} = W\bar{h}}Z^{d\bar{g}_t}\EV \big[Z^{\bar{h}_t}Z^{ \tilde{h}}\big] \label{eqn:main_zdeltag}\\
    \text{where}\quad Z^{W\delta\tilde{h}_{t+1}} &= \hat{Z}^{W\delta\tilde{h}_{t+1}} + \sum_y Z^{\tilde{y}}\EV \f{\partial Z^{\delta\tilde{h}_{t+1}}} {\partial \hat Z^{W^\trsp \tilde{y}}}.
\end{align}
\item If $g = \psi(h^1,...,h^k)$, then using \cref{eqn:delta_nonlin,eqn:psistar}, taking the limit $1/\sqrt n \to 0$,
\begin{align}\label{eqn:nonlinp}
    Z^{\delta\tilde{g}_{t+1}} = \sum_{i=1}^k \frac{\partial \psi(Z^{\tilde{h}^1},...,Z^{\tilde{h}^k})}{\partial Z^{\tilde{h}^i}}Z^{\delta\tilde{h}^i_{t+1}}.
\end{align}
\item Using \cref{eqn:g_update,eqn:dupfate1,eqn:dupfate2,eqn:dupfate3} and taking $\frac{1}{\sqrt{n}} \to 0$, we have by \refZNonlinp{}:
\begin{equation}
    Z^{g_t(\xi)} = Z^{g_0(\xi)},~~~Z^{dg_t(\xi)} = Z^{dg_0(\xi)}\quad\text{for any vector $g \in \pi_0$ at time $t$}.\label{eqn:Zxt}
\end{equation}
\end{itemize}

\subsection{Deriving The NTK}

Instantiate paths $p$ and $q$ on two inputs $\xi, \xi'$ by $p=p(\xi), q=q(\xi')$ (abusing notation slightly).
We define an inner product between them as follows:
\begin{align}\label{eqn:path_product}
\big\langle p , q\big\rangle \defeq \EV \big[ Z^{p^0}Z^{q^0}\big]\prod_{i=2,even}^{|p|-2}\EV\big[\frac{\partial Z^{p^i}}{\partial Z^{p^{i-1}}}\frac{\partial Z^{q^i}}{\partial Z^{q^{i-1}}}\big].
\end{align}
where $p^i,q^i$ are X-vars for all even $i$.
Note that for even $i$, $p^i$ is always of the form $p^i = \psi(....,p^{i-1},...)$ for some $\psi$. So the partial derivatives in \cref{eqn:path_product} are just $\frac{\partial Z^{p^i}}{\partial Z^{p^{i-1}(\tilde{\xi})}} = \psi'(...,Z^{p^{i-1}},...)$.

Our goal in this section is to prove
\begin{prop}\label{prop:NTK_paths}
    \begin{align}
        \mathring{\mathcal{K}}(\xi,\tilde{\xi}) = \sum_{p, q:p^{-1} = q^{-1} = x,p \cong q}\langle p({\xi}), q(\tilde \xi) \rangle. \label{eqn:NTK_paths}
     \end{align}
\end{prop}
For each weight $W \in \mathcal{W}$, the gradient of the output with respect to $w$ is given by:
\begin{align}
    \nabla_w f(\xi) = \sum_{g,h:g = Wh}\frac{dg(\xi)\ h(\xi)^\top}{n}
\end{align}
Here, $g,h$ represent nodes in program $\pi_0$ that can be instantiated by an input $\xi$.\\
The NTK of $f$ can be expressed as:
\begin{align}
    \mathring{\mathcal{K}}(\xi,\tilde{\xi}) &= \lim_{n \to \infty}\sum_{W \in \mathcal{W}}\big\langle \nabla_w f(\xi) ,\nabla_w f(\tilde{\xi})\big\rangle\\
    &= \lim_{n \to \infty}\sum_{W \in \mathcal{W}} \sum_{g,h:g = Wh}\sum_{\bar{g},\bar{h}:\bar{g} = W\bar{h}}\frac{dg(\xi)^\top d\bar{g}(\tilde{\xi})}{n}\frac{h(\xi)^\top \bar{h}(\tilde{\xi})}{n}\\
    &= \sum_{W \in \mathcal{W}} \sum_{g,h:g = Wh}\sum_{\bar{g},\bar{h}:\bar{g} = W\bar{h}}\EV \big[Z^{dg(\xi)}Z^{d\bar{g}(\tilde{\xi})}\big]\EV \big[Z^{h(\xi)}Z^{\bar{h}(\tilde{\xi})}\big]. \label{eqn:paths_eq}
\end{align}

Using \cref{eqn:error+paths,eqn:path_der}, for any G-var $g =Wh$, we can write the error term $dg$ as the summation of errors signals over paths $p$: 
\begin{align}
dg(\xi) =  \sum_{p:p^{-1} = x,p^{1} = g} J^p(\xi),~~~~~~~J^p = \psi'(...,p^1,...) \odot \big((W^{p^3})^\top J^{p:3}\big)
\end{align}
Hence we can write:
\begin{align}
    \EV  \big[Z^{dg(\xi)}Z^{dg(\tilde{\xi})}\big]
    &= \sum_{p^1=g,p^{-1}=x,q^1=g,q^{-1} = x}\EV \big[Z^{J^p(\xi)}Z^{J^q(\tilde{\xi})}\big]\label{eqn:error_sum}\\
    Z^{J^p } &= Z^{(W^{p^3})^\top J^{p:3}} \frac{\partial{Z^{p^2}}}{\partial Z^{p^1}}
\end{align}
where $\psi'$ denotes the derivative w.r.t.\ $p^1$.
By Simple GIA Check \citep{yang2}, we have that $Z^{(W^{p^3})^\top J^{p:3}} = \hat{Z}^{(W^{p^3})^\top J^{p:3}}$ (see \refZMatMul{}).
Hence, with abuse of notation $J^p = J^p(\xi),J^q = J^q(\tilde{\xi}), p = p(\xi), q = q(\tilde \xi)$, we have
\begin{align}
    \EV \big[Z^{J^p}Z^{J^q}\big] = \EV[\hat{Z}^{(W^{p^3})^\top J^{p:3}}\hat{Z}^{(W^{q^3})^\top J^{q:3}}]\EV[\frac{\partial{Z^{p^2}}}{\partial Z^{p^1}}\frac{\partial{Z^{q^2}}}{\partial Z^{q^1}}].
\end{align}
From the definition of \refZhat{}, the expectation $\EV[\hat{Z}^{(W^{p^3})^\top J^{p:3}}\hat{Z}^{(W^{q^3})^\top J^{q:3}}]$ vanishes if the weights $W^{p^3}$ and $W^{q^3}$ are not symbolically the same (i.e $W^{p^3}  \triangleq W^{q^3}$).
Then by \refZhat{},
\begin{align}\label{eqn:recusive}
    \EV[\hat{Z}^{(W^{p^3})^\top J^{p:3}}\hat{Z}^{`(W^{q^3})^\top J^{q:3}}] =
    \begin{cases}
        \EV[Z^{ J^{p:3}}Z^{ J^{q:3}}] & \text{if $W^{p^3} \triangleq W^{q^3}$}\\
        0 & \text{otherwise.}
    \end{cases}
\end{align}
Applying this logic recursively, we have
\begin{align}
    \EV \big[Z^{J^p}Z^{J^q}\big] = 
    \begin{cases}
        \prod_{i=2,even}^{|p|-2}\EV\big[\frac{\partial Z^{p^i}}{\partial Z^{p^{i-1}}}\frac{\partial Z^{q^i}}{\partial Z^{q^{i-1}}}\big]
            & \text{if $p \cong q$}\\
        0 & \text{otherwise.}
    \end{cases}
\end{align}

Combining with \cref{eqn:paths_eq,eqn:error+paths,eqn:error_sum} proves \cref{prop:NTK_paths}.

\subsection{Getting \cref{claim:NTK}}
\paragraph{Notation}
For the remainder of the proof we abbreviate $p = p(\tilde{\xi}), q = q(\xi_t)$, $p^i = p^i(\tilde{\xi}), q^i = q^i(\xi_t)$ (i.e path $p$ is always evaluated on $\tilde{\xi}$, while path $q$ is always evaluated on $\xi_t$).\\
We prove \cref{claim:NTK} by inducting on all G-vars in the network. We begin by proving the following induction hypothesis.

\begin{defn}
We write $Z^{x}\equiv Z^{y}\mod\hat{Z}^{W\bullet}$ to denote that $Z^{x}-Z^{y}$ is a linear combination of $\hat{Z}^{Wu}$ for various vectors $u$.
\end{defn}

\newtheorem*{indhyp}{Induction Hypothesis}
\begin{indhyp}
At any time $t$ and G-var $g=Wh$, the following holds:
\begin{equation}
Z^{\delta \tilde{g}_{t+1}}\equiv -\chi_t\sum_{p:p^{-1}=g}\sum_{q:q\cong p}Z^{dq^{-1}}\langle p,q\rangle\mod\hat{Z}^{W\bullet}\label{eq:IH}
\end{equation}
\end{indhyp}
Here, the sum is over all paths $p$ with endpoint $g$ and all paths $q$ isomorphic to $p$.
Recall that $dq^{-1}$ is the (scaled) gradient $dy$ where $y = q^{-1}$ is the endpoint of $q$.

\subsubsection{Base Case}

For initial G-vars $g$, $\delta \tilde{g}_t=0$ since we are not training the input layers (Assumption \ref{ass:first_last}). This proves the base case since the sum in \cref{eq:IH} has no terms and thus is 0.

\subsubsection{Inductive case}

Suppose $g=Wh$, where $h=\psi(h^1,...,h^k)$, we then have using \cref{eqn:main_zdeltag}:
\begin{align}
Z^{\delta \tilde{g}_{t+1}} & \equiv\dot{Z}^{W\delta \tilde{h}_{t+1}}+\chi_t\sum_{\bar{g}=W\bar{h}}Z^{d\bar{g}_t}\EV \big[ Z^{\bar{h}_t}Z^{\tilde{h}}\big] \mod\hat{Z}^{W\bullet}\label{eqn:Zdeltatildeg}\\
\text{where}\quad\dot{Z}^{W\delta \tilde{h}_{t+1}} & =\sum_{y}Z^{y}\EV\frac{\partial Z^{\delta \tilde{h}_{t+1}}}{\partial\hat{Z}^{W^{\trsp}y}}.\label{eqn:ZdotWdeltah}
\end{align}
Note $\sum_{\bar{g}=W\bar{h}}Z^{d\bar{g}_t}\EV \big[ Z^{\bar{h}_t}Z^{\tilde{h}}\big]$ in \cref{eqn:Zdeltatildeg} can be written as $\sum_{\bar{p}:\bar{p}^{-1} = g,|\bar{p}|=2}\sum_{\bar{q}\cong \bar{p}}Z^{d\bar{q}^{-1}}\langle \bar{p}, \bar{q}\rangle$.
Therefore, it suffices to show that
\begin{equation}
    \dot{Z}^{W\delta \tilde{h}_{t+1}}
    = -\chi_t\sum_{\bar{p}:\bar{p}^{-1} = g,|\bar{p}|\geq 4}\sum_{\bar{q}\cong \bar{p}}Z^{d\bar{q}^{-1}}\langle \bar{p}, \bar{q}\rangle.\label{eqn:ZdotWdeltatildeh}
\end{equation}
\paragraph{Showing \cref{eqn:ZdotWdeltatildeh}}
By \cref{eqn:delta_nonlin}: 
\begin{align}\label{eqn:zdeltah_main}
Z^{\delta \tilde{h}_{t+1}}=\sum_{i=1}^k \frac{\partial Z^{\tilde{h}}}{\partial Z^{\tilde{h}^i}}Z^{\delta\tilde{h}^i_{t+1}}.
\end{align}
Since $Z^{\tilde{h}^1},...,Z^{\tilde{h}^k}$ do not depend on $Z^{W^{\trsp}y}$ for any $y$ (by the assumption that we don't use both a matrix and its transpose in the forward pass), from \cref{eqn:zdeltah_main} we have for any $y$:
\begin{align}\label{eqn:no_trans1}
\EV\big[\frac{\partial Z^{\delta \tilde{h}_{t+1}}}{\partial\hat{Z}^{W^{\trsp}y}}\big] = \sum_{i=1}^k \EV\big[\frac{\partial Z^{\tilde{h}}}{\partial Z^{\tilde{h}^i}}\frac{\partial Z^{\delta\tilde{h}^i_{t+1}}}{\partial \hat{Z}^{W^\top y}}\big].
\end{align}
Applying the induction hypothesis \cref{eq:IH} to each G-var $h^i$, we get
\begin{equation}
\frac{\partial Z^{\delta\tilde{h}^i_{t+1}}}{\partial \hat{Z}^{W^\top y}} = -\chi_t\sum_{p:p^{-1}=h^i}\sum_{q:q\cong p}\frac{\partial Z^{dq^{-1}}}{\partial \hat{Z}^{W^\top y}}\langle p,q\rangle\mod\hat{Z}^{W\bullet} \label{eqn:dDeltaG/dW^T}
\end{equation}

Plugging this back into $\dot{Z}^{W\delta \tilde{h}_{t+1}}$ (\cref{eqn:ZdotWdeltah}), we get
\begin{align}
\dot{Z}^{W\delta \tilde{h}_{t+1}} = -\chi_t\sum_{y}Z^{y}\sum_{i=1}^{k}\sum_{p:p^{-1}=h^i}\sum_{q:q\cong p}\langle p,q\rangle\EV\big[\frac{\partial Z^{\tilde{h}}}{\partial Z^{\tilde{h}^i}}\frac{\partial Z^{dq^{-1}}}{\partial \hat{Z}^{W^\top y}}\big].\label{eqn:ZWDeltaH}
\end{align}

Note that for any path $p$ with $p^{-1} = h^i$, we may extend $p$ by vectors $g,h$ (recall $g = Wh$ and $h = \psi(h^1,...,h^k)$).
Let $\bar p$ denote this extension.
If $\bar q$ is a path such that
\begin{equation}
    \bar q \cong \bar p \quad\text{and}\quad \frac{\partial Z^{dq^{-1}}}{\partial \hat{Z}^{W^\top y}} = \frac{\partial Z^{\bar{q}^{-2}}}{\partial Z^{\bar{q}^{-3}}},\label{eqn:condq}
\end{equation}
then
\begin{align}
    \langle p,q\rangle\EV\big[\frac{\partial Z^{\tilde{h}}}{\partial Z^{\tilde{h}^i}}\frac{\partial Z^{dq^{-1}}}{\partial \hat{Z}^{W^\top y}}\big] = \langle \bar{p},\bar{q}\rangle.
\end{align}
Our goal now is to show $q$ in \cref{eqn:ZWDeltaH} can be extended appropriately such that we may rewrite \cref{eqn:ZWDeltaH} as \cref{eqn:ZdotWdeltatildeh}.
This will be done through explicitly computing the term $\frac{\partial Z^{dq^{-1}}}{\partial \hat{Z}^{W^\top y}}$ in \cref{eqn:ZWDeltaH}.

\paragraph{Computing $\frac{\partial Z^{dq^{-1}}}{\partial \hat{Z}^{W^\top y}}$}
Suppose $\{g^1,...,g^r\}$ are all G-vars in the program $\pi_0$ that depend on $q^{-1}$ i.e for all $1\leq j \leq r,$ we have $g^j = W^{j}z^j$ where $z^j = \psi_j (...,q^{-1},...)$ and where $W^j$ can be same or different matrices for different $j$.
Note that it follows that:
\begin{align}
    dq^{-1} &= \sum_{j=1}^r \psi_j' \big(...,q^{-1},...\big) \odot ((W^{j})^\top dg^j)\\
    Z^{dq^{-1}} &= \sum_{j=1}^r \frac{\partial Z^{z^j}}{\partial Z^{q^{-1}}}Z^{(W^{j})^\top dg^j}\label{eqn:Zdq-1}\\
    \text{where}\quad Z^{(W^{j})^\top dg^j} &= \hat{Z}^{(W^{j})^\top dg^j} + \dot{Z}^{(W^{j})^\top dg^j} = \hat{Z}^{(W^{j})^\top dg^j} \label{eqn:GIA_check}.
\end{align}
Note that $\dot{Z}^{(W^{j})^\top dg^j} = 0$ in \cref{eqn:GIA_check} from the gradient independence assumption (GIA) because we pass the Simple GIA Check.
This may also be easily verified by explicitly computing $\dot{Z}^{(W^{j})^\top dg^j} = \sum_y Z^y\EV \frac{\partial Z^{dg^j}}{\partial \hat{Z}^{Wy}}$, and noticing that the expectation vanishes from the dependency of $Z^{dg^j}$ on $Z^v$ (i.e $Z^{dg^j} = Z^vZ^\mu$ for some vector $\mu$ which does not depend on $v$). Since $y$ does not depend on $v$ and the last layer $v$ is not trained, we have $\EV \frac{\partial Z^{dg^j}}{\partial \hat{Z}^{Wy}} = \EV [Z^v] \EV[..] = 0$. \\
Since we assumed that the forward propagation does not contain both $W,W^\top$, it follows from differentiating \cref{eqn:Zdq-1} that
\begin{align}\label{eqn:no_trans2}
    \frac{\partial Z^{dq^{-1}}}{\partial \hat{Z}^{W^\top y}} 
    =
        \sum_{j=1}^r \frac{\partial Z^{z^j}}{\partial Z^{q^{-1}}}\frac{\partial\hat{Z}^{(W^{j})^\top dg^j}}{\partial \hat{Z}^{W^\top y}}
    =
        \sum_{j:W^j \triangleq W,dg^j = y} \frac{\partial Z^{z^j}}{\partial Z^{q^{-1}}}. 
\end{align}
If this sum over $j$ is nonempty, then there is a unique $j$ such that $W^j \triangleq W$ and $dg^j = y$.
In such a case, we may extend the path $q$ with $g^j,z^j$ to form $\bar{q}$ satisfying \cref{eqn:condq}.
Plugging back into \cref{eqn:ZWDeltaH} we obtain \cref{eqn:ZdotWdeltatildeh} as desired.

Hence, we have proven the induction hypothesis.
\subsubsection{Proving \cref{claim:NTK} using the induction hypothesis}
WLOG assume $f(\xi) = V^\top x(\xi)$ for some G-var $x(\xi)$.
Using the induction hypothesis and the Master Theorem (\cref{thm:PLNetsorT+MasterTheorem}), we have that:
\begin{align}
    \lim_{n \to \infty} \tilde{f}_{t+1} -  \tilde{f}_t = \EV Z^v Z^{\delta\tilde{x}_{t+1}} = -\chi_t\sum_{p:p^{-1} = x}\sum_{q\cong q}\EV \big[Z^vZ^{dq^{-1}}\big]\langle p, q\rangle .
\end{align}
Note that $Z^{dq^{-1}} = Z^v$ for any path $q:q^{-1} = x$. Hence, with \cref{eqn:NTK_paths}, we have
\begin{align}
     \lim_{n \to \infty} \tilde{f}_{t+1} -  \tilde{f}_t =-\chi_t\sum_{p:p^{-1} = x}\sum_{q\cong p}\langle p, q\rangle
     =-\chi_t \mathring{\mathcal{K}}(\xi_t,\tilde{\xi})
\end{align}
as desired.

\crefname{assumptionsi}{Assumption}{Assumptions}
\crefrangeformat{assumptionsi}{Assumptions (#3#1#4) to (#5#2#6)}

\subsection{Relaxing \cref{ass:first_last,ass:trans,ass:scalar,ass:GvarEmbedding}}

We now briefly discuss the case where \cref{ass:first_last,ass:trans,ass:scalar,ass:GvarEmbedding} are relaxed, as well as the case where $f$ is represented by a \netsortp{} program. As the proof of the general case follows roughly the same logic as in \cref{setup:main_simplified}, we only discuss the meaningful differences in each case.
\subsubsection{Training the first and last layers}
Recall the input and output layers are parameterized by $\{u^i\},v$ which now depend on $t$. The output evolution is now given by:
\begin{align}\label{eqn:output_ass1}
    \tilde{f}_{t+1} - \tilde{f}_t = V_{t+1}^\top\tilde{x}_{t+1} - V_t^\top\tilde{x}_t = \frac{v^\top \delta \tilde{x}_{t+1}}{n} + \frac{\sqrt{n}(v_{t+1} - v_t)^\top \tilde{x}_t}{n} + \sum_{s=0}^t\frac{(v_{s+1} - v_s)^\top \delta\tilde{x}_{t+1}}{n}.
\end{align}
Plugging $v_{t+1} - v_t = -\chi_t \frac{1}{\sqrt{n}}x_t$ into \cref{eqn:output_ass1} and taking the limit (using \netsortp{} rules):
\[
\lim_{n \to \infty} \tilde{f}_{t+1} - \tilde{f}_t = \EV\big[Z^v Z^{\delta \tilde{x}_{t+1}} \big] -\chi_t \EV\big[Z^{x(\xi_t)}Z^{\tilde{x}}\big].
\]
Evaluating $\EV\big[Z^v Z^{\delta \tilde{x}_{t+1}} \big]$ by induction requires altering the path definition so that each path $p$ may start with an input $\xi$, and ends with a G-var (that is, a path either starts with an X-var or an input). We reuse the definition of inner product between $p \cong q$ in \cref{eqn:path_product}, only when both start with inputs $\xi,\tilde{\xi}$ respectively then $\EV\big[Z^{p^0}Z^{q^0}\big]$ implies $\xi^\top \tilde{\xi}$. The remainder of the proof follows the same logic as with \cref{setup:main_simplified}. Note that the NTK in this case would yield:
\[
\mathring{\mathcal{K}}(\xi_t,\tilde{\xi}) = \sum_{q\cong q}\langle p, q\rangle + \EV\big[Z^{x(\xi_t)}Z^{\tilde{x}}\big].
\]
\subsubsection{$W,W^\top$ in the forward pass}
When both $W,W^\top$ are allowed in the forward pass, the update equations for each $w_t$ take the form:
\[
w_{t+1} - w_t = -\chi_t \sum_{g,h:g=Wh}\frac{dg_t h_t^\top}{n} - \chi_t \sum_{g,h:g=W^\top h}\frac{h_t dg_t^\top }{n}
\]
Some quick calculations using \netsortp rules show that for G-vars:
\begin{itemize}
    \item If $g = Wh$:
    \[
    Z^{\delta \tilde{g}_{t+1}} = Z^{W\delta \tilde{h}_{t+1}} - \chi_t \sum_{\bar{g},\bar{h}:\bar{g} = W\bar{h}}Z^{d\bar{g}(\xi_t)}\EV\big[Z^{\bar{h}(\xi_t)}Z^{\tilde{h}}\big]) - \chi_t \sum_{\bar{g},\bar{h}:\bar{g} = W^\top\bar{h}}Z^{\bar{h}(\xi_t)}\EV\big[Z^{d\bar{g}(\xi_t)}Z^{\tilde{h}}\big].
    \]
    \item If $g = W^\top h$:
    \[
    Z^{\delta \tilde{g}_{t+1}} = Z^{W^\top\delta \tilde{h}_{t+1}} - \chi_t \sum_{\bar{g},\bar{h}:\bar{g} = W^\top\bar{h}}Z^{d\bar{g}(\xi_t)}\EV\big[Z^{\bar{h}(\xi_t)}Z^{\tilde{h}}\big]) - \chi_t \sum_{\bar{g},\bar{h}:\bar{g} = W\bar{h}}Z^{\bar{h}(\xi_t)}\EV\big[Z^{d\bar{g}(\xi_t)}Z^{\tilde{h}}\big].
    \]
\end{itemize}
It is straightforward to show using GIA \citep{yang2} that $\EV\big[Z^{d\bar{g}(\xi_t)}Z^{\tilde{h}}\big] = 0$ in both cases, leaving us with a similar expression as with \cref{setup:main_simplified}. The induction hypothesis for G-vars in this case takes one of two forms:
\begin{itemize}
    \item If $g = Wh$ then \cref{eq:IH} holds.
    \item If $g = W^\top h$ then \cref{eq:IH} holds with $\mod\hat{Z}^{W^\top\bullet}$ replacing $\mod\hat{Z}^{W\bullet}$.
\end{itemize}

Some additional complications need to be resolved. Specifically, with setup \cref{setup:main_simplified} we have used in two places the fact that no transpose is used in the forward pass to prove the induction hypothesis (see \cref{eqn:no_trans1,eqn:no_trans2}). 
To prove the induction, and assuming $g = Wh$, we now have instead of \cref{eqn:no_trans1} (using \cref{eqn:zdeltah_main}):
\begin{align}\label{eqn:with_trans1}
\EV\frac{\partial Z^{\delta \tilde{h}_{t+1}}}{\partial\hat{Z}^{W^{\trsp}y}} = \sum_{i=1}^k \EV\big[\frac{\partial Z^{\tilde{h}}}{\partial Z^{\tilde{h}^i}}\frac{\partial Z^{\delta\tilde{h}^i_{t+1}}}{\partial \hat{Z}^{W^\top y}}\big] + \sum_{i=1}^k \EV\big[\frac{\partial^2 Z^{\tilde{h}}}{\partial Z^{\tilde{h}^i}\partial \hat{Z}^{W^\top y}}Z^{\delta\tilde{h}^i_{t+1}}\big].
\end{align}
where $\{h^i\}$ are G-vars. To evaluate the additional term on the RHS of \cref{eqn:with_trans1}, we use the induction hypothesis to express $Z^{\delta\tilde{h}^i_{t+1}}$:
\begin{align}\label{eqn:vanishing_exp}
    \EV\big[\frac{\partial^2 Z^{\tilde{h}}}{\partial Z^{\tilde{h}^i}\partial \hat{Z}^{W^\top y}}Z^{\delta\tilde{h}^i_{t+1}}\big] = - \chi_t \sum_{p:p^{-1}=h^i}\sum_{q\cong p}\langle p, q\rangle \EV\big[\frac{\partial^2 Z^{\tilde{h}}}{\partial Z^{\tilde{h}^i}\partial \hat{Z}^{W^\top y}} Z^{dq^{-1}}\big].
\end{align}
Using GIA \cite{yang2}, it is straight forward to show that the expectation on the RHS of \cref{eqn:vanishing_exp} vanishes, leaving us with the first term on the RHS of \cref{eqn:with_trans1}, as with \cref{setup:main_simplified}. Note that the same logic may be applied in \cref{eqn:no_trans2}, concluding the proof.

\subsubsection{Multiple outputs and arbitrary batchsize}
We have used a scalar output and a batchsize of 1 throughout this paper. However, extending to multiple (finite) outputs and an arbitrary batchsize requires no additional arguments besides some additional notations. For example, the definition of path should now be altered to express dependency on multiple samples (if batchnorm is used for example). The proof however follows roughly the same logic in \cref{setup:main_simplified}.

\subsubsection{X-var embedding}
We assumed in our proof that $x$, which represents the final embedding of $f$ is a G-var. However, extending the proof to the case where $x$ is an X-var is straightforward. Let $f(\xi) = V^\top x(\xi)$ where $x = \psi(h^1,...,h^k)$ and $h^1,...,h^k$ are G-vars. Using the induction hypothesis, along with \cref{eqn:nonlinp} yields:
\begin{align}
        \lim_{n \to \infty} \tilde{f}_{t+1} -  \tilde{f}_t &=-\chi_t\EV\big[Z^vZ^{\delta\tilde{x}_{t+1}}\big] = -\chi_t\sum_{i=1}^k\EV\big[Z^v \frac{\partial Z^{\tilde{x}}}{\partial Z^{\tilde{h}^i}}Z^{\delta\tilde{h}_{t+1}}\big]\\
        &=-\chi_t\sum_{i=1}^k\EV\big[Z^v \frac{\partial Z^{\tilde{x}}}{\partial Z^{\tilde{h}^i}}\sum_{p:p^{-1} = h^i}\sum_{q\cong p}Z^{dh^i(\xi_t)}\langle p,q \rangle\big]\\
        &=-\chi_t\sum_{i=1}^k \sum_{p:p^{-1} = h^i}\sum_{q\cong p}\langle p,q \rangle\EV\big[\frac{\partial Z^{\tilde{x}}}{\partial Z^{\tilde{h}^i}}\frac{\partial Z^{x(\xi_t)}}{\partial Z^{h^i(\xi_t)}}\big] \label{eqn:ntknonlin}\\
        &= -\chi_t \mathring{\mathcal{K}}(\xi_t,\tilde{\xi})
\end{align}
It is straightforward to show that the expression for $\mathring{\mathcal{K}}(\xi_t,\tilde{\xi})$ in \cref{eqn:ntknonlin} represents the NTK if this case.

\subsubsection{Network specified by \netsortp}

If the network is more generally represented by a \netsortp{} program instead of just a \netsort{} program, then our proof can be very simply modified to accommodate as follows:
The new operation allowed in such a network is the production of a scalar through \refMoment{}, say $a = \f 1 n \sum_{\alpha=1}^n \psi(x^1_\alpha, \ldots, x^k_\alpha; \theta^1, \ldots, \theta^l)$.
By a similar inductive argument as before, we will see that
1) $x^i_t = x^i_0 + o(1)$ for all $i \in [k]$ and $\theta^j_t = \theta^j_0 + o(1)$ for all $j \in [l]$, so that $a_t = a_0 + o(1)$;
2) in the backward pass, any backpropagation through $a$ will zero out: For example, if $a$ is only used later in a \refNonlin{} $z = \bar \psi (y; a)$, then $\f 1 {\sqrt n} \nabla_a f = \langle dz, \partial_a \bar \psi(y; a) \rangle/n$ will converge to 0 because of GIA (as $dz$ is linear in the final layer), and the error signal at $x^i$ times $\sqrt n$ is the constant vector with entries $\f 1 {\sqrt n} \nabla_a f$, which is $o(1)$.

Therefore, we can treat any scalar produced through \refMoment{} as a constant fixed at initialization, and the notion of path from before carries over here without change (by assuming all nonlinearities with scalar parameters to be parameterless nonlinearities where the parameters are fixed).
Then the same reasoning follows.

\end{document}